\documentclass[10pt,journal,compsoc]{IEEEtran}

\usepackage[pdftex]{graphicx}

\usepackage{spconf,amsmath}
\usepackage{amssymb}
\usepackage{booktabs}
\usepackage{multirow}
\usepackage{xcolor}

\ifCLASSOPTIONcaptionsoff
 \usepackage[nomarkers]{endfloat}
\let\MYoriglatexcaption\caption
\renewcommand{\caption}[2][\relax]{\MYoriglatexcaption[#2]{#2}}
\fi

\ifCLASSOPTIONcompsoc
  \usepackage[nocompress]{cite}
\else
  \usepackage{cite}
\fi

\newcommand\ans[1]{\textcolor{black}{#1}}
\title{ISP Distillation}
\name{Eli Schwartz$^{\dagger}$ \thanks{Corresponding author: Eli Schwartz \texttt{me@eli-schwartz.com}}
\qquad Alex M. Bronstein$^{\ddagger}$ \qquad Raja Giryes$^{\dagger}$ }
  
  \address{$^{\dagger}$ Tel Aviv University \qquad
      $^{\ddagger}$ Technion -- Israel Institute of Technology}

\begin{document}
%\ninept
%
\maketitle
%
%%%%%%%%% ABSTRACT

\begin{abstract}
Nowadays, many of the images captured are `observed' by machines only and not by humans, e.g., in autonomous systems.
High-level machine vision models, such as object recognition or semantic segmentation, assume images are transformed into some canonical image space by the camera \ans{Image Signal Processor (ISP)}.
However, the camera ISP is optimized for producing visually pleasing images for human observers and not for machines. Therefore, one may spare the ISP compute time and apply vision models directly to RAW images.
Yet, it has been shown that training such models directly on RAW images results in a performance drop.
To mitigate this drop, we use a RAW and RGB image pairs dataset, which can be easily acquired with no human labeling. We then train a model that is applied directly to the RAW data by using knowledge distillation such that the model predictions for RAW images will be aligned with the predictions of an off-the-shelf pre-trained model for processed RGB images.
Our experiments show that our performance on RAW images for object classification and semantic segmentation is significantly better than models trained on labeled RAW images.
It also reasonably matches the predictions of a pre-trained model on processed RGB images, while saving the ISP compute overhead.
\end{abstract}

%

%%%%%%%%% BODY TEXT

% \begin{figure*}[h!]
%     \centering
%     \def\svgwidth{\textwidth}
%     \includegraphics[width=1\textwidth]{figures/concept-fig-horiz.pdf}
%     \vspace{-10pt}
%     \caption{ISP Distillation: Distilling the knowledge of the ISP algorithm together with the CNN classifier that follows it improve the performance of a classifier applied directly on the RAW images. Optionally, the classifier can operate on short-exposure RAW images, effectively distilling the information in the physical process of acquiring a signal with better SNR.}
%     \label{fig:concept}

% \end{figure*}

\begin{figure}
    \centering
    \includegraphics[width=1.01\linewidth]{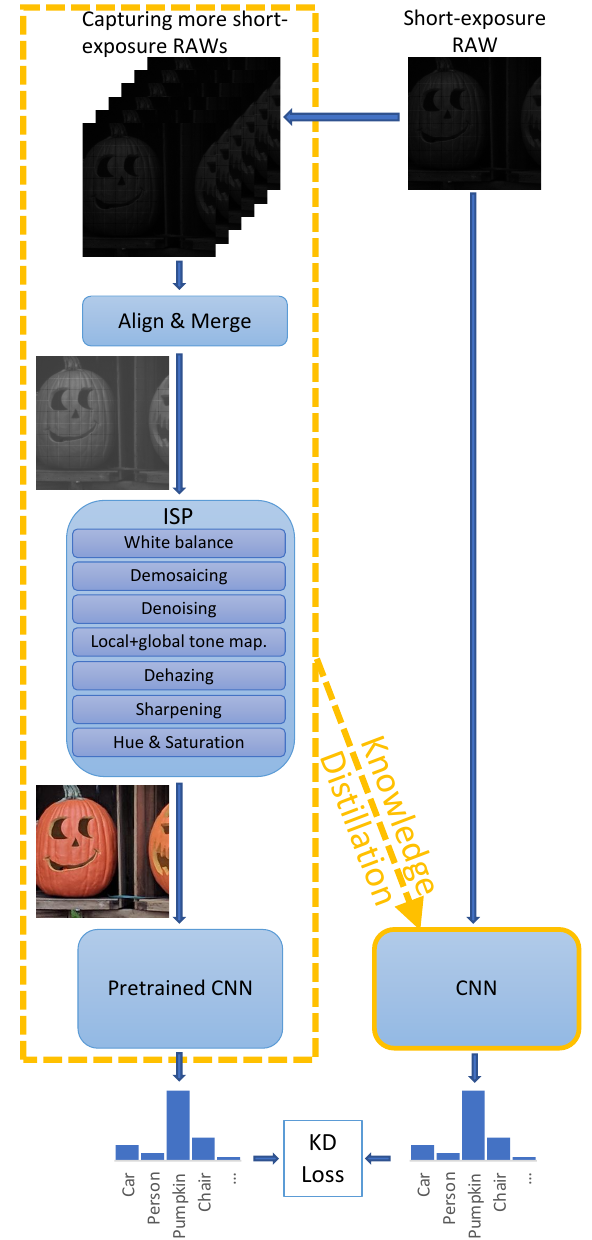}
    \vspace{-10pt}
    \caption{ISP Distillation: Distilling the knowledge of the ISP algorithm together with the pretrained CNN (e.g. classifier or semantic segmentation) that follows it improves the performance of a CNN applied directly to RAW images. Optionally, the CNN can operate on short-exposure RAW images, effectively distilling the information in the physical process of acquiring a signal with better SNR.}
    \label{fig:concept}
\end{figure}

\begin{figure*}
    \centering
    \begin{tabular}{ccc}
        Input & Output segmentation mask & Mask overlaid on input \\
         \includegraphics[width=0.31\linewidth]{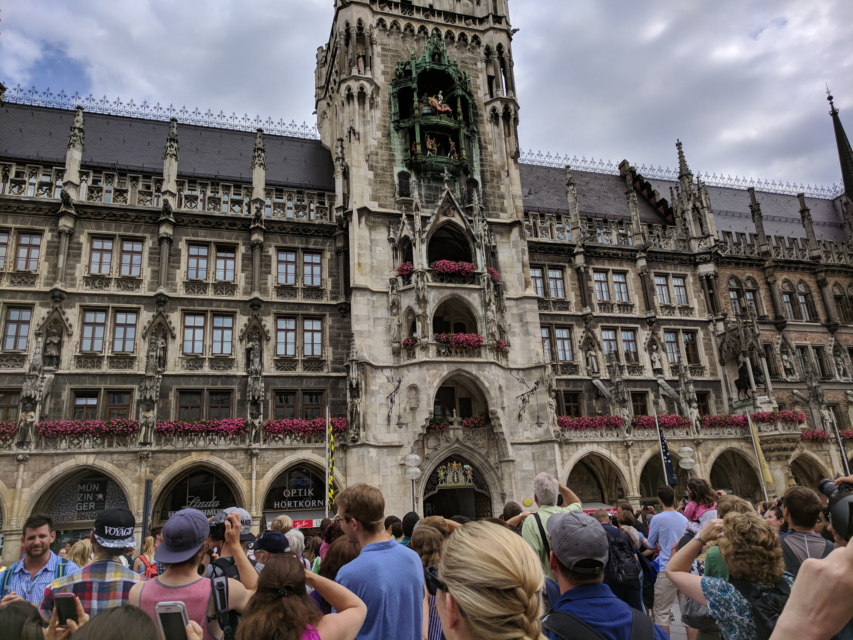} & \includegraphics[width=0.31\linewidth]{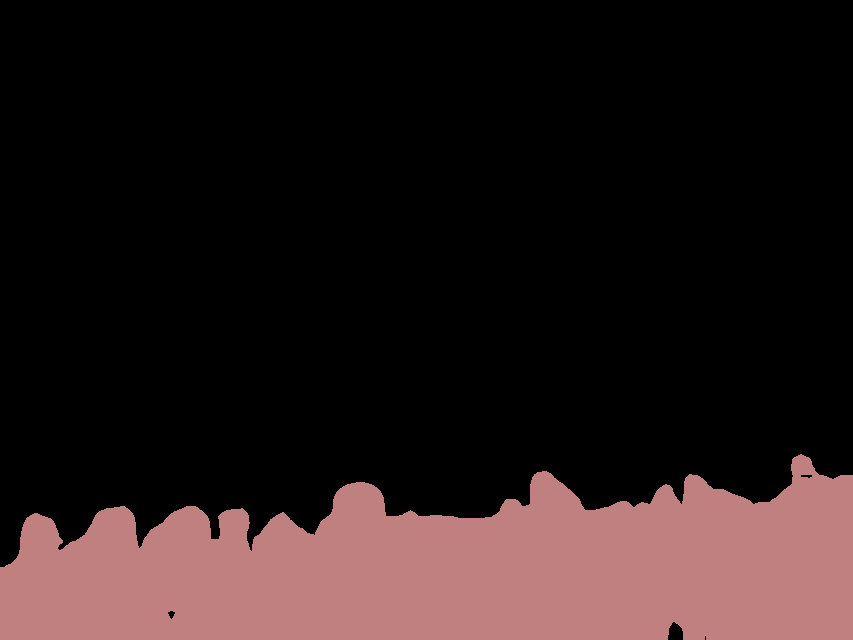} & \includegraphics[width=0.31\linewidth]{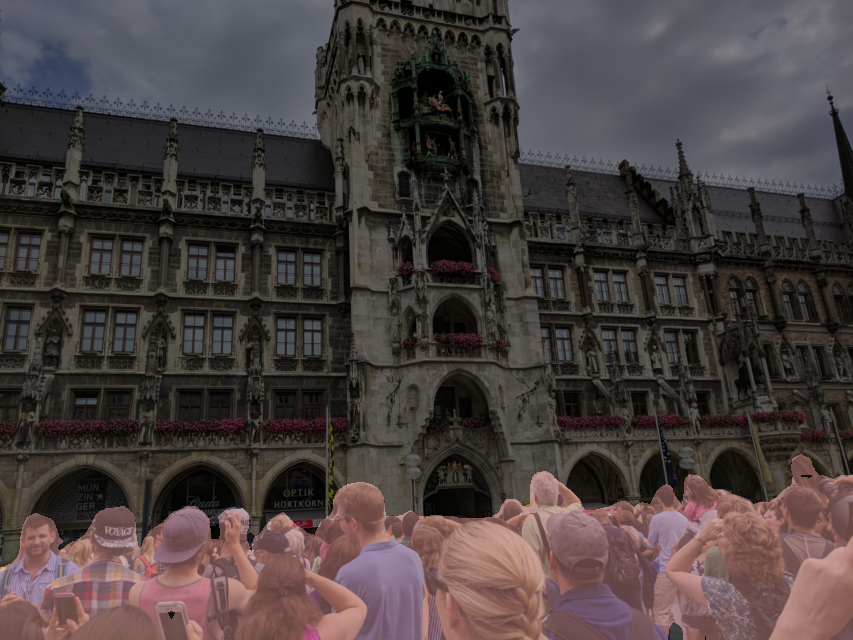} \\
         \includegraphics[width=0.31\linewidth]{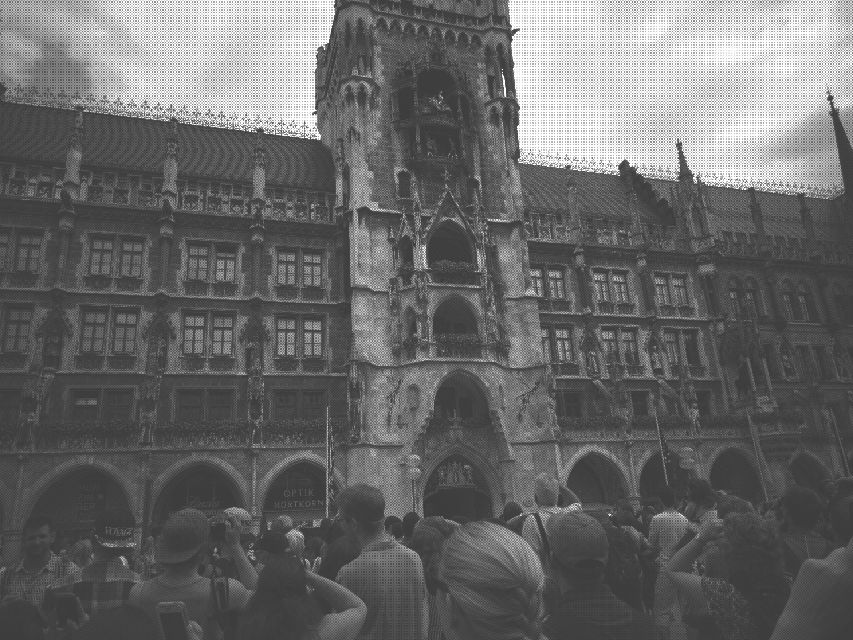} & \includegraphics[width=0.31\linewidth]{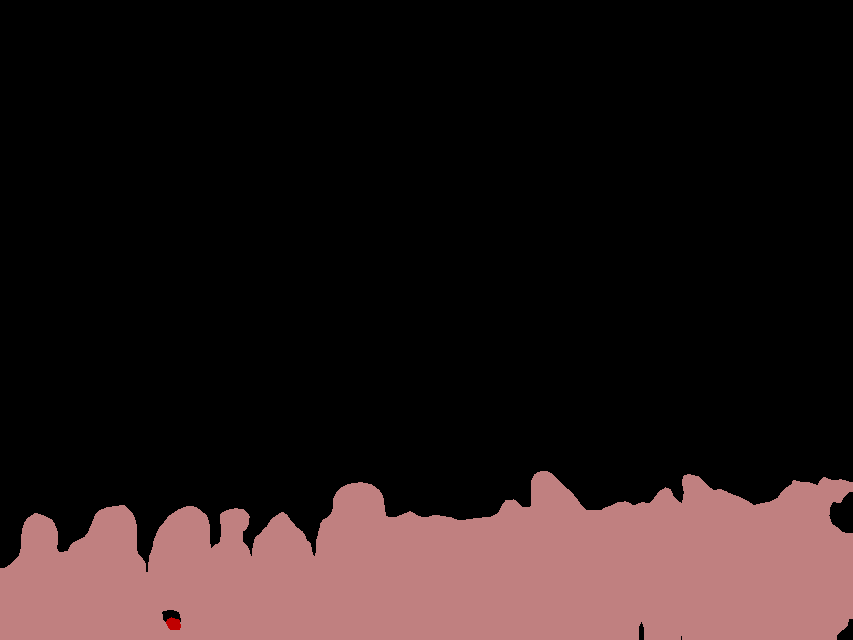} & \includegraphics[width=0.31\linewidth]{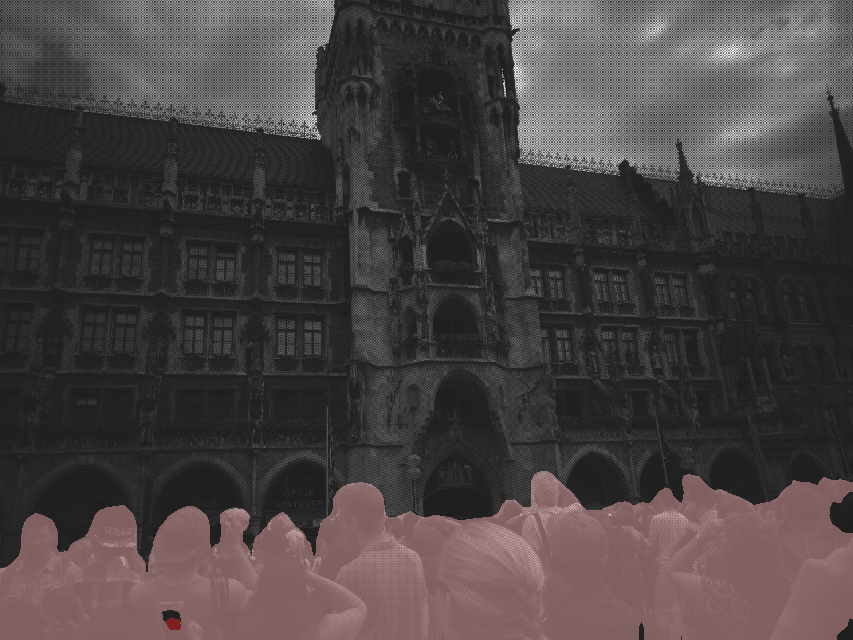} \\
         \includegraphics[width=0.31\linewidth]{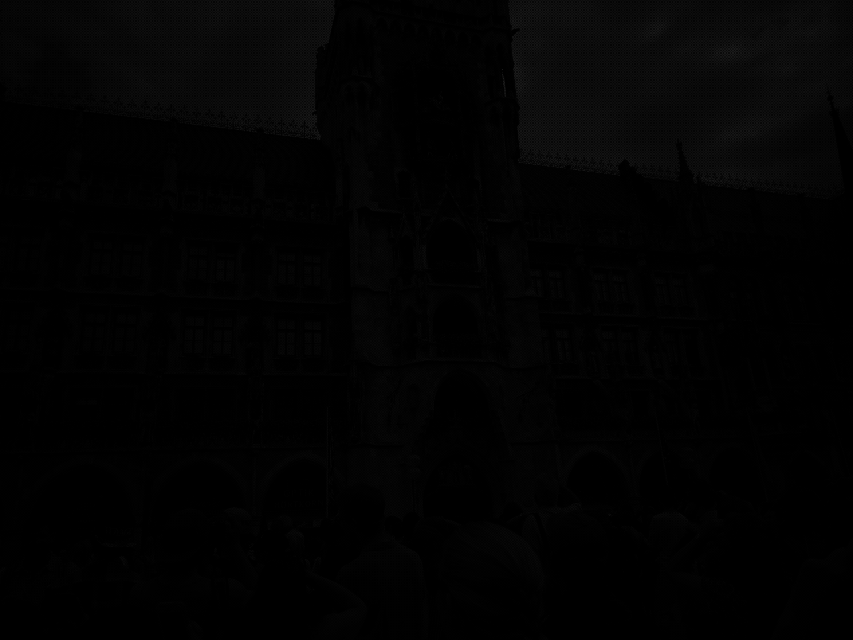} & \includegraphics[width=0.31\linewidth]{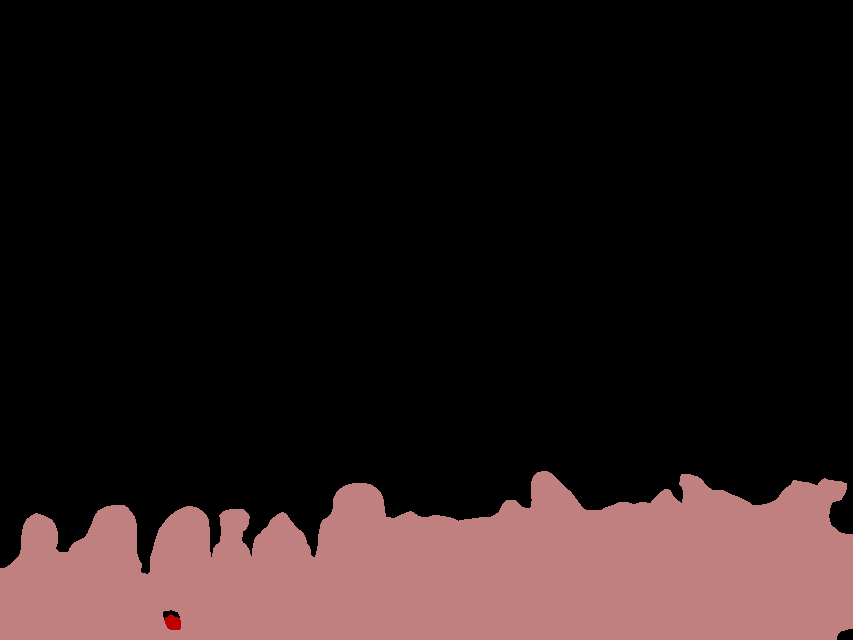} & \includegraphics[width=0.31\linewidth]{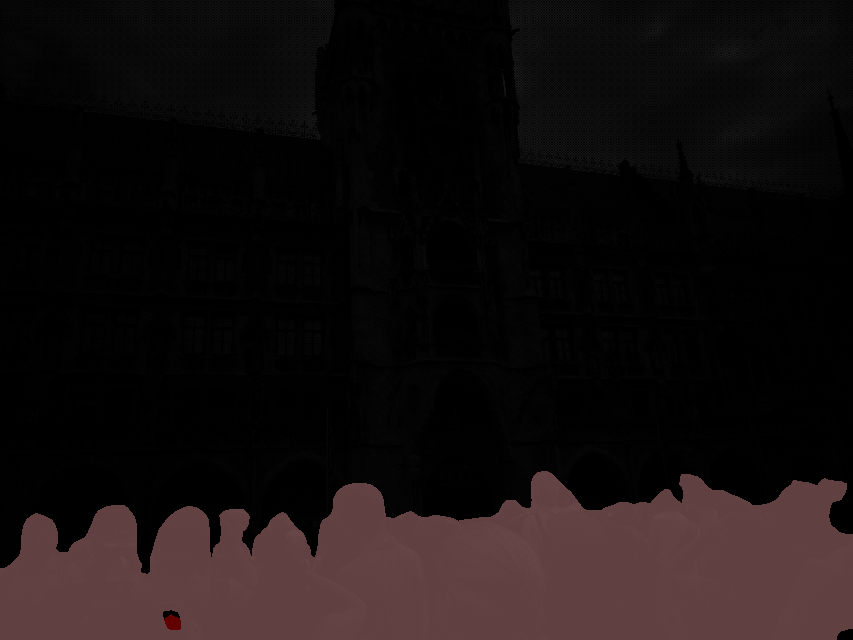} \\
    \end{tabular}
    \caption{\textbf{Semantic segmentation result example:} Top row is the ground truth RGB results; middle row is our results for RAW input images; bottom row is our results for short-exposure RAW images.}
    \label{fig:seg1}
\end{figure*}

\section{Introduction}
Traditionally, Image Signal Processors (ISPs) are designed to optimize human \ans{perception} of photographs. The core task is, given a raw measurement from an array of sensors (with different sensitivity to different light frequencies), to produce an image that looks natural to the human observer. To this end, there is a need to estimate the right colors and tones, and also compensate for acquisition process artifacts, such as noise.

However, in many domains, scenes are captured for machine consumption only. Examples include robot cameras, autonomous driving, and security cameras that are automatically monitored. In these domains, the objective is not to produce visually pleasing images, but rather to achieve high performance in a given downstream task, e.g. object recognition. Thus, discarding the ISP and training the model directly on the raw data is tempting in these cases. 
\ans{An ISP is typically a dedicated hardware system designed and optimized for low latency. Discarding it can reduce design costs, silicon area, and the power used.}
Unfortunately, simply discarding the ISP \ans{has been} shown in the literature to cause performance degradation \ans{for high-level computer vision tasks such as object-recognition} \cite{hansen2019isp4ml,diamond2017dirty,buckler2017reconfiguring}. Note though that for lower-level tasks such as optical-flow it might be beneficial \cite{zheng2020optical}). This happens because the ISP serves as a `normalization' of the data, transforming it into a canonical space, which is independent (or less dependent) of the camera used or the capturing environment. Solutions discussed in the literature usually involve adjustments of the ISP for the vision task, either manually \cite{yahiaoui2019overview} or via learning \cite{wu2019visionisp}. 
\ans{
Unlike alternative methods that focus on designing a minimal ISP for vision, in this paper we focus on completely discarding the ISP. We explore how the drop in performance can be mitigated.
}

Training a model on RAW images requires annotating the data. Human labeling of RAW images is impossible, as they are sometimes almost unrecognizable to humans. The alternative is to use RGB images for labeling and then transfer the labels to RAW. The transfer to RAW can be done by having an inverse model of the ISP ($ISP_{inv}$). So given a dataset of labeled RGB images $\{(RGB_i,l_i)\}$, we can generate a matching labeled RAW dataset $\{(RAW_i,l_i)\}=\{(ISP_{inv}(RGB_i),l_i)\}$
relying on pixel alignment between RAW and RGB to get the labels for the RAW \cite{diamond2017dirty,hansen2019isp4ml,buckler2017reconfiguring}. 
%But the ISP is not invertible and thus the simulated inverse model is not accurate and due to the one-to-many mapping should be stochastic. Also, designing the inverse model is not trivial.

In this work, instead of using an ISP inverse model for transferring labels from RGB to RAW, we employ a dataset of RAW-RGB pairs. Building such a dataset is relatively easy since it does not require human labeling and by labeling the RGB images we immediately get labels for the RAW. Also, we wish to reduce the labeling cost. Thus, instead of manually labeling the RGB images we use a pre-trained model to label them.

With the above-mentioned dataset we can train our model on RAW images with the transferred labels as ground truth. Yet, we still suffer from a big drop in performance compared to working with RGB inputs. To mitigate that, we suggest using Knowledge Distillation (KD) \cite{hinton2015distilling} to make the RAW predictions fit the RGB predictions. KD is a technique that is known to work quite well for compressing deep models, i.e., making a smaller model behave similarly to a larger model. Here we use KD for compressing both the heuristically designed ISP \textit{and} a higher-level deep vision model, e.g. classifier, 
into a new deep model for classification but with the size of only the latter part. We show the advantage of our approach also when training the network to have similar predictions for short-exposure RAW images. In that case, in addition to the ISP and classification, the model also compresses the `knowledge' of an ideal (non-existent) denoiser that maps the short-exposure RAW to a longer-exposure RAW image. We show that this technique works well for both classification and semantic segmentation.

\section{Related Work}

% \subsection{ISP}

% \subsubsection{Traditional ISP}

% \subsubsection{Learned ISP}

\noindent \textbf{ISP for Vision.}
The ISP consists of a set of algorithms, usually applied sequentially, intended for transforming a RAW image into a visually appealing RGB image. The different steps either fix some degradation in the acquisition process (e.g., noise) or just transform the image to better fit \ans{human perception} (e.g., tone mapping). Recently, some works suggested replacing some or all of these operations with a learned model \cite{schwartz2018deepisp,chen2018learning}. However, these designed or learned ISPs are optimized for visual appearance and not the vision tasks.

Simply dropping the ISP does not work well. Several works used simulated RAW images to train a classifier and observed a substantial gap in accuracy \cite{diamond2017dirty}.
Hansen et al. \cite{hansen2019isp4ml} report a larger gap for smaller models,  $\sim$$16\%$ drop for MobileNet, attributed to the failure of compact models to compensate for the lack of ISP.
Buckler et al. \cite{buckler2017reconfiguring} identified the lack of demosaicing and gamma correction to be a critical cause of performance degradation.
They suggested modifying the imaging sensor such that demosaicing and gamma correction are no longer necessary, which limits the effect of the lack of ISP.

Some methods suggest optimizing the ISP for downstream vision tasks.
\ans{\cite{yahiaoui2019overview,tseng2019hyperparameter,mosleh2020hardware} suggested tuning the parameters of a traditional (not learned) ISP to improve downstream tasks.}
Sharma et al. \cite{sharma2018classification} add a component that takes an RGB image processed by the ISP and further enhances it for the downstream task. 
Diamond et al. \cite{diamond2017dirty} suggested jointly learning a low-level processing module, which performs denoising and deblurring, with the classifier.
They train with simulated RAW images.
Wu et al. \cite{wu2019visionisp} suggested VisionISP, a trainable ISP, which is trained to optimize object detection in an autonomous driving setting.
Wang et al. \cite{wang2020deep} suggested a totally different approach for performing classification on low-quality images (e.g. fog or low-contrast). They use a model pre-trained on high-quality images and learned a mapping between the deep representations of the low-quality images and the high-quality images. 
\ans{Essentially, unlike other methods, they perform domain adaptation 
by mapping the outputs of the classifier rather than the inputs.
Other methods for classifying low-quality images include \cite{8396981,borkar2019deepcorrect,zheng2016improving}.}
In this work, instead of designing an image processing module, i.e., modifying the input image for a certain task, we focus on applying the vision model directly to RAW data.

% \subsection{Domain Adaptation}

\noindent \textbf{Knowledge Distillation.}
Compressing larger models into smaller ones was first suggested by Bucilua et al. \cite{bucilua2006model}.
Application of this technique to deep neural networks, known as Knowledge Distillation, was suggested by Hinton et al. \cite{hinton2015distilling}.
The key idea behind KD is that the soft label output (or soft probabilities) of a classifier contains much more information about the data point than the hard label.
KD has been widely used for numerous different applications, e.g. \cite{bagherinezhad2018label, li2017learning, peng2019few}. It has also been shown to work for distilling knowledge of non-neural network machine-learning models \cite{fukui2019distilling}. In \cite{chi2020dynamic,gnanasambandam2020image} KD was used to transfer knowledge across different types of sensors for either image reconstruction or high-level tasks, and in \cite{hong2021student} it was theoretically analyzed.
In this work, we use KD to distill not just a deep model, but also the non-learned (manually engineered) algorithms in the ISP and the information gained in the physical process of acquiring a better signal (that has a better SNR).

\section{Method}
\label{sec:method}
% \vspace{-10pt}
We are interested in training a classification model to operate on images from one modality, when no semantic labels are available.
What we have is a function that maps the images to another modality for which we have a pre-trained model.
Alternatively, we might have a dataset of image pairs from two different modalities with pixel alignment, even if a function that maps these modalities does not exist (or we do not have access to it).
In our case, the two modalities are a $RAW$ and a processed image $RGB=ISP(RAW)$.
We are also interested in operating on a short exposure image, $RAW_{short}$, where the reference processed image is based on a longer exposure, $RGB_{long}=ISP(RAW_{long})$, or multiple short exposure RAW images, $RGB_{long}=ISP(\{RAW_{short}^i\})$ (clearly, there is no deterministic mapping $f$ such that $RGB_{long}=f(RAW_{short})$).

Manually labeling RAW images is hard as they are almost impossible to be recognized by humans. The alternative is to use RGB images for labeling and transfer the labels to RAW. The transferring to RAW can be done by having an inverse model of the ISP ($ISP_{inv}$). So given a dataset of labeled RGB images $\{(RGB_i,l_i)\}$, we can generate a matching labeled RAW dataset $\{(RAW_i,l_i)\}=\{(ISP_{inv}(RGB_i),l_i)\}$
relying on pixel alignment between RAW and RGB to get the labels for the RAW. But the ISP is not invertible and thus the inverse model is just an estimation and due to the one-to-many mapping should be stochastic. Also, designing the inverse model is not trivial.
In this work, instead of having an inverse model of the ISP for transferring the labels from RGB to RAW, we use a dataset of RAW-RGB pairs. Moreover, to reduce the labeling cost, instead of manually labeling the RGB images, we use a pretrained model, $M_{RGB}$, to label them.

\begin{table}[tb]
    \centering
    \normalsize	
    % \hspace{-10pt}
    \begin{tabular}{lcccc}
        \toprule
        & \multicolumn{2}{c}{ResNet18} & \multicolumn{2}{c}{MobileNetV2} \\
                                                    & top-1 & top-5 & top-1 & top-5 \\
         \midrule
         Clean images \footnotesize{(upper bound)}                 & 69.76 & 89.08 & 71.88 & 90.29 \\
         \midrule
         \textbf{Baselines}: \\
         Pretrained model                           & 29.23 & 52.53 & 25.65 & 48.73 \\
         Trained w/ GT labels                     & 57.21 & 80.76 & 56.31 & 80.06 \\
         Trained w/ pred. labels  & 56.59 & 79.70 & 56.73 & 80.15 \\
         \midrule
        %  ISP Distillation (ours)                   & \textbf{61.35} & \textbf{83.70} & \textbf{61.43} & \textbf{83.95}\\
        %  ISP Distillation (teacher ResNet50)                   & \textbf{62.43} & \textbf{84.18} & \textbf{62.62} & \textbf{84.14}\\
         ISP Distillation (ours) & \textbf{62.46} & \textbf{84.41} & \textbf{62.92} & \textbf{84.84} \\ 
        %  Gradual Distillation from clean image      &       &       \\
         \bottomrule
    \end{tabular}
% \vspace{-10pt}
    \caption{Classification performance on noisy and mosaiced images. We use the ImageNet images in a Color Filter Array format with added synthetic Gaussian noise with STD=0.1.}
% \vspace{-15pt}
    \label{tab:imagenet}
\end{table}

We argue that training on the dataset $\{(RAW_i,l_i)\}$, where $l_i$ are the hard labels (e.g., an integer value or a one-hot encoding), with the cross-entropy (CE) loss is not the best option. Instead of using $l_i$, we use the soft probability distribution vector over the classes, $p_i$, predicted by a pretrained model. 
Following many works that have shown its benefits, we use the KD loss. 

Given the probability vectors 
$p = M_{RGB}(RGB)$ and $q = M_{RAW}(RAW)$, which are the outputs of a softmax layer with temperature $T$.
\ans{
A softamx with a temperature $T$ for a logit vector $x$ is defined as
\begin{equation} 
p_i = \frac{e^{x_{i}/T}}{\sum^{K}_{k=1}e^{x_{k}/T}}.
\end{equation}
The temperature $T$ controls the ``softness" of the predicted distribution. For smaller $T$ we get a low-entropy vector (an approximation of one-hot), while for higher $T$ we get a high-entropy vector (the probability is distributed more evenly between all classes).
}
The KD loss is given by 
\ans{
\begin{equation}
L_{KD}=-\sum{p_i \cdot\log(q_i)}.
\end{equation}
}
This loss simultaneously distills the information from the heuristically designed ISP and the CNN model (classifier) pretrained on RGB images, $M_{RGB}$. 

An extension of the KD loss that can be beneficial is adding an $\ell_2$ loss on the intermediate representation vectors \ans{\cite{romero2014fitnets}}. We found this extension to be useful in our case too. We use the extended form of the loss
\ans{
\begin{equation}
L_{KD}=\ell_2(f_{RGB},f_{RAW}) - \sum{p_i \cdot\log(q_i)},
\end{equation}
}
where $f_{RGB}$ and $f_{RAW}$ are the intermediate representations of the RGB image from $M_{RGB}$ and the RAW image from $M_{RAW}$. 
\ans{
Unlike \cite{romero2014fitnets} where the representation is captured in the middle of the network, we use the representation from the last layer before the classifier. This is more likely to produce aligned probability predictions, even without balancing the CE loss and the features loss magnitudes. 
}

In the case of short-exposure RAW, we use 
\begin{align}
&p=M_{RGB}(RGB_{long}) \\
&q=M_{RAW}(RAW_{short}),
\end{align}
and we also implicitly learn to classify low-dynamic-range and extremely noisy images. As commonly done, we use a linear combination of the Cross-Entropy (CE) loss and KD loss
\begin{equation}
    L=\alpha L_{CE} + (1-\alpha) L_{KD},
\end{equation}
\ans{
Where the CE loss is defined as
\begin{equation}
    L_{CE} = -\sum{p_i \log(y_i)}
\end{equation}
($y$ is the ground-truth one-hot probability vector).
In our experiments, we chose $T=4$ (similar to \cite{hinton2015distilling}) and we heuristically chose $\alpha=0.9$ to balance the two loss terms (same order of magnitude).
}

We also extend the method to the semantic segmentation task. Where the class label (or class probabilities) is predicted at each pixel or spatial location. In this case, the loss is averaged over all spatial locations at the output layer (which might be different from the input resolution). 
\ans{
In the case of spatial elements equal to $N$ and $L_{CE}(i), L_{KD}(i)$ the loss functions at location $i$, the final loss is defined as:
\begin{equation}
    L=\frac{1}{N} \sum_{i=1}^{N}{\alpha L_{CE}(i) + (1-\alpha) L_{KD}(i)}.
\end{equation}
}

It is common practice to normalize the inputs so each channel has zero-mean and unit STD (calculated on the relevant training set). For the RAW images, we do the same, where the mean and STD are calculated separately for R, G, and B pixels in the Bayer pattern.

In practice, for faster training, we initialize the RAW classifier, $M_{RAW}$, with the weights of the pre-trained RGB model $M_{RGB}$. Using this initialization, we can resort to a short training (4 epochs in our experiments). 
Since RAW images have a single channel and not 3 RGB channels, we need to make some adaptations to be able to use off-the-shelf classifiers with them (especially when initializing with pre-trained models). We do so in a very simple way by transforming the RAW images to RGB and filling in the missing values using bilinear interpolation, similar to what is done in \cite{schwartz2018deepisp}. 
\ans{There is no loss of information in bilinear interpolation. The original pixel values are unchanged, and the interpolation only fills in the missing values introduced by the conversion from the 1-channel Bayer pattern to RGB.
}

\begin{figure}[t]
    \centering
    \includegraphics[width=1\linewidth]{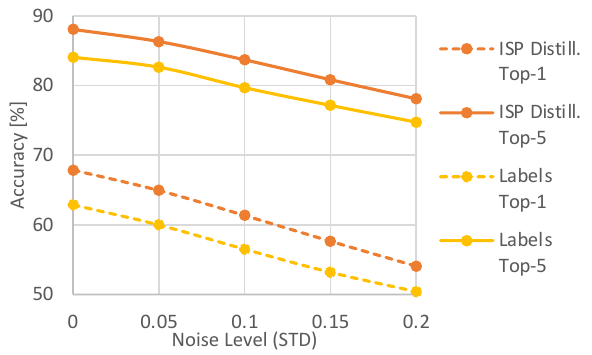}
% \vspace{-10pt}
    \caption{Performance vs. noise level for noisy mosaiced ImageNet with ResNet18.}
% \vspace{-10pt}
    \label{fig:acc-vs-noise}
\end{figure}

\section{Experiments}
% \vspace{-10pt}
\ans{To validate our approach we used two test cases.} In the first, we test performance when operating on noisy and mosaiced images, i.e., discarding the denoising and demosaicing pre-processing. In the second, we test performance when the full ISP is discarded.

% \vspace{-10pt}
\subsection{Discarding Denoising and Demosaicing}

We first test our method on synthetically generated RAW images, since for these images we can compare to the classification performance of training with ground truth labels. In this experiment, we limit the simulated ISP (we want to discard) to include only denoising and demosaicing. Thus, the RAW images are the RGB images sub-sampled according to a Bayer pattern with added Gaussian noise. We use the ImageNet dataset for this experiment (ILSVRC2017 \cite{russakovsky2015imagenet}).

Table~\ref{tab:imagenet} presents ResNet18 \cite{he2016deep} and MobileNetV2 \cite{sandler2018mobilenetv2} results for the mosaiced and noisy ImageNet validation set, with noise STD equals to $0.1$ (pixel values in $[0,1]$). Under such a distortion, the performance of ResNet18 drops from $69.76\%$ top-1 accuracy to $29.23\%$. Training the model on distorted images (RAW) using either ground truth labels or hard labels produced by a pretrained ResNet18 (based on the clean RGB images) improves performance to $\sim 57\%$. Using the proposed ISP Distillation we improve the performance by more than $4\%$. Similar trends are observed for MobileNetV2 too. Fig.~\ref{fig:acc-vs-noise} shows the improvement is consistent across noise levels, including for $STD=0$ where it is just demosaicing.

\begin{table}[tb]
    \centering
    \normalsize	
    % \hspace{-10pt}
    \begin{tabular}{@{}llrrrr@{}}
         \toprule
         && \multicolumn{2}{c}{ResNet18} &  \multicolumn{2}{c}{MobileNet V2} \\
                                                    && top-1 & top-5 & top-1 & top-5 \\
        \midrule
         \multicolumn{5}{l}{\textbf{Classifying RAW images:}} \\
         & Training with labels     & 84.73  & 95.40 & 87.16 & 97.37 \\
         & ISP Distillation  (ours)            & \textbf{93.83}  & \textbf{98.12}  & \textbf{93.32} & \textbf{98.08}\\
         \midrule
         \midrule
         \multicolumn{5}{l}{\textbf{Classifying \textit{short-exposure} RAW images:}} \\
         & Training with labels     & 83.73  & 95.31  &  80.62 & 94.48     \\
         & ISP Distillation (ours)             & \textbf{91.87}  & \textbf{96.98}  & \textbf{91.57} & \textbf{96.99} \\
         \cmidrule(l{1em}r{1em}){2-6}
        %  \cmidrule(){2-6}
         & With ISP (DeepISP \cite{schwartz2018deepisp}) & 91.86 & 97.67 & 92.30 & 97.71 \\
         \bottomrule
    \end{tabular}
    % \vspace{3pt}
    \caption{Performance when discarding the full ISP and training the classifier on RAW images (or short-exposure RAW). \ans{Measuring the agreement between the RAW model and pre-trained RGB model predictions on an automatically collected test set.} ISP Distillation consistently outperforms training with labels (labels predicted on RGB). Note that ISP distillation achieves competitive results to the computationally demanding solution of using an ISP network \cite{schwartz2018deepisp} prior to the classifier.}
% \vspace{-10pt}
    \label{tab:raw}
\end{table}

\begin{table}[tb]
    \centering
    \normalsize	
    % \hspace{-10pt}
    \ans{
    \begin{tabular}{@{}llcccc@{}}
         \toprule
         && \multicolumn{2}{c}{ResNet18} &  \multicolumn{2}{c}{MobileNet V2} \\
                                                    && top-1 & top-5 & top-1 & top-5 \\
        \midrule
         \multicolumn{5}{l}{\textbf{Classifying RAW images:}} \\
         & Training with labels     & 82.93  & 93.83 & 83.88 & 94.78 \\
         & ISP Distillation  (ours) & \textbf{90.99}  & \textbf{94.78}  & \textbf{90.99} & \textbf{96.20}\\
         \midrule
         \midrule
         \multicolumn{5}{l}{\textbf{Classifying \textit{short-exposure} RAW images:}} \\
         & Training with labels     & 81.99  & 92.89  &  80.09 & 91.46     \\
         & ISP Distillation (ours) & \textbf{90.04}  & \textbf{94.78}  & \textbf{89.09} & \textbf{95.26} \\
         \bottomrule
    \end{tabular}
    }
    % \vspace{3pt}
    \caption{\ans{Performance when discarding the full ISP and training the classifier on RAW images (or short-exposure RAW) measured on a manually labeled test set (211 images). ISP Distillation consistently outperforms training with labels (labels predicted on RGB).}}
% \vspace{-10pt}
    \label{tab:raw_labled_test}
\end{table}

\begin{table}[h]
    \centering
    \normalsize	
    % \hspace{-10pt}
    \begin{tabular}{lcrr}
         \toprule
           & FLOPs & top-1 & top-5   \\
        \midrule
        DeepISP + ResNet18 & 4.5G & 91.86 & 97.67  \\
         ResNet18 & 2G  & 91.87 & 96.98 \\
       \midrule
        DeepISP + MobileNetV2 & 2.8G & 92.30 & 97.71  \\
         MobileNetV2 & 0.3G  & 91.57  & 96.99  \\
         \bottomrule
    \end{tabular}
    % \vspace{3pt}
    \caption{\textbf{Accuracy vs. FLOPs.} Our classifier-only model reach comparable results to training an ISP and a classifier while having much lower computation cost (FLOPs).}
% \vspace{-10pt}
    \label{tab:compute}
\end{table}

\begin{table}[h]
    \centering
    \normalsize	
    % \hspace{-10pt}
    \begin{tabular}{p{0pt}lc}
         \toprule
         && mIOU \\
        \midrule
         \multicolumn{2}{l}{\textbf{RAW images:}} \\
         & Training with labels     & 92.47\% \\
         & ISP Distillation  (ours)            & \textbf{94.98}\%  \\
         \midrule
         \midrule
         \multicolumn{2}{l}{\textbf{\textit{short-exposure} RAW images:}} \\
         & Training with labels     & 91.88\% \\
         & ISP Distillation (ours)             & \textbf{94.51}\% \\
         \bottomrule
    \end{tabular}
    % \vspace{3pt}
    \caption{\textbf{Semantic segmentation performance}. mIOU is computed as agreement with the pretrained model predictions for the corresponding RGB image.}
% \vspace{-10pt}
    \label{tab:segmentation}
\end{table}

\begin{table}[h]
    \centering
    \normalsize	
    \begin{tabular}{lc}
        \toprule
                                                    & top-1 \\
         \midrule
         Training the first quarter of the model   & 90.61 \\
         Training the first half of the model   & 91.66 \\
         Training Full model            & \textbf{91.87} \\
         \bottomrule
    \end{tabular}
    % \vspace{3pt}
    \caption{Finetuning only the first layers. Most of the gain is due to finetuning the first and second quarters of the model. Tested with ResNet18 on short-exposure RAW.}
% \vspace{-15pt}
    \label{tab:ablation}
\end{table}

\subsection{Discarding the Full ISP}
\label{sec:disc-isp}
% \vspace{-5pt}
To test the effect of discarding the full ISP, we use real data based on images from the HDR+ dataset \cite{hasinoff2016burst}. This dataset includes $3640$ bursts (containing $28461$ images in total). Each burst has between 2 to 10 short-exposure raw photos, where each is generally 12-13 Mpixels, depending on the type of camera used for the capturing. The images in a burst are generally captured with the same exposure time and gain. The dataset also provides for each burst a merged RAW image that is generated by aligning the short-exposure RAW images and combining them to produce a single high-dynamic-range RAW. This merged RAW is then processed by their ISP to produce the final RGB image. See \cite{hasinoff2016burst} for more details.

We performed two kinds of experiments. In the first, we want to distill the ISP and a pre-trained classification model. Thus, we use the merged RAW as input to our model, trying to mimic the predictions on the final RGB. In the second experiment, we choose a single short-exposure RAW and train our model to mimic the predictions \ans{made by the RGB model} on the final RGB, \textit{produced from the merged RAW}. We always choose the short-exposure RAW that is pixel-aligned with the merged RAW (information provided in the dataset).

\ans{For evaluation we used both a larger pseudo-label dataset and a smaller manually labeled dataset. For the pseudo-labeled dataset} the top-1 and top-5 accuracies are measured as the agreement of our model with the predictions of the pre-trained classification model on the final RGB. The bursts are randomly split into $80\%$ training set and $20\%$ test set. The images in the original HDR+ dataset are of very high resolution, but popular classifier architectures expect images to be in the range of $200-300$ pixels. While we could down-sample the images, it would have removed the effect of the Bayer pattern (and potentially the effect of the noise too), and we are interested in understating the ability of the model to overcome these artifacts. Therefore, we chose to split the images into smaller $256\times256$ non-overlapping patches. For training, we use all the patches originating from the bursts in the training set. For the test, since many of the patches do not contain an object, we only tested those for which the pre-trained RGB classifier predicted an object with probability $p>0.8$.
\ans{The manually labeled test set contains 211 images out of the pseudo-labeled dataset test set described above.}

Table~\ref{tab:raw} compares the performance of training the RAW model with predicted labels vs. applying ISP Distillation. Our approach shows a substantial improvement of $+9\%$ top-1 accuracy for ResNet18 and $+6\%$ for MobileNet V2. It also exhibits a similar advantage for the model trained on short-exposure RAW, where the improvement is $+8\%$ and $+11\%$ for ResNet18 and MobileNet V2, respectively. A noticeable advantage exists in all experiments for top-5 accuracy too, bringing the accuracy to $97-98\%$. This suggests that the RAW model predicted probability distribution highly matches the one from the RGB model. 
\ans{Table~\ref{tab:raw_labled_test} compares the performance on the small manually labeled test set. While the results are a bit lower compared to the pseudo-labeled test set, we observe similar trends.}

%DeepISP is able to process a short-exposure RAW image to look like a long-exposure RGB. 
For the short-exposure RAW experiment, we also compared to a sequential combination of a network that performs the ISP part (DeepISP \cite{schwartz2018deepisp}) before the classifier, where both are trained end-to-end to optimize classification performance. The computational costs vs. accuracy are summarized in Table \ref{tab:compute}.
DeepISP flop count is $\sim2.5G$ (compared to ResNet18's $2G$ and MobileNetV2's $0.3G$).
Note that the classifier alone, trained with ISP Distillation, performs almost as well as the combination of models, which is more computationally demanding. The expressive power of the joint ISP and classifier is probably limited by the constraint of having a 3-channel tensor at their interface.  When training just the classifier, using our approach, both models are mostly compressed into the classifier alone.

\subsection{Semantic segmentation}

We tested the extension to semantic segmentation, where the KD loss is applied at each spatial location, as mentioned in Sec. \ref{sec:method}. For this experiment we used the DeepLabV3 model \cite{chen2017rethinking} with ResNet101 backbone \cite{he2016deep}. The model used for the RGB images was trained on Pascal VOC dataset \cite{Everingham10} and a subset of the COCO dataset \cite{lin2014microsoft} with images containing objects from the 20 classes in the VOC dataset. As in the classification case, the same pretrained weights were used to initialize the model operating on RAW images.

\begin{figure*}[h!]
    \centering
    \begin{tabular}{ccc}
        Input & Output segmentation mask & Mask overlaid on input \\
         \includegraphics[width=0.25\linewidth]{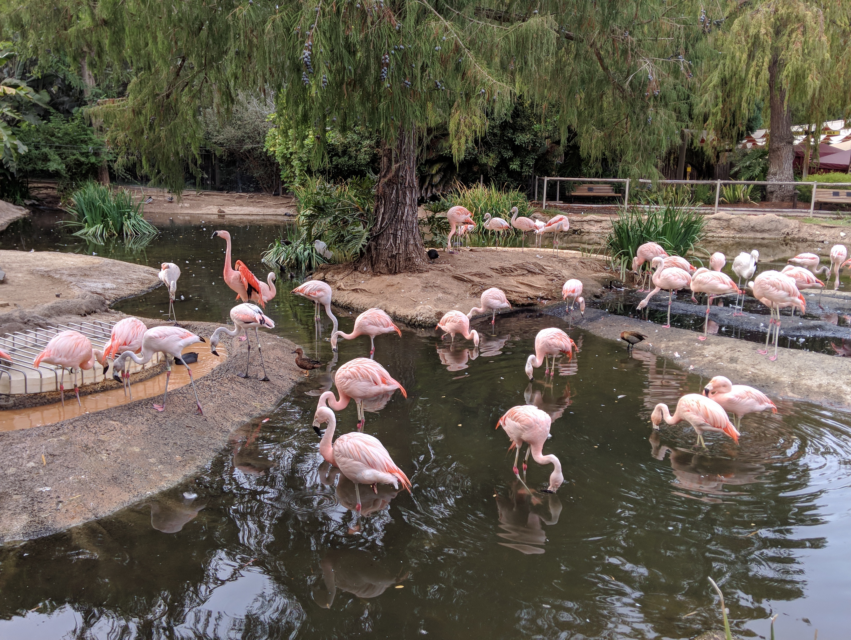} & \includegraphics[width=0.25\linewidth]{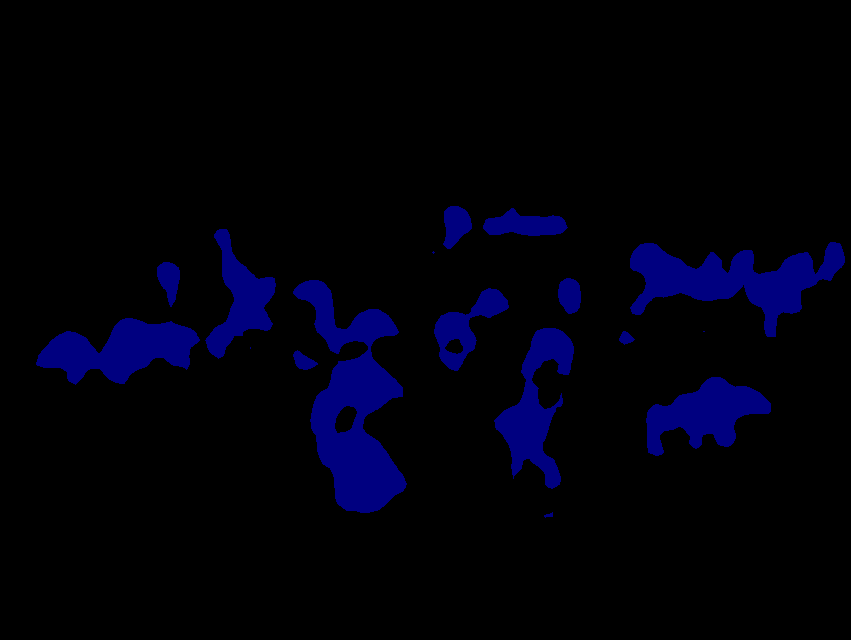} & \includegraphics[width=0.25\linewidth]{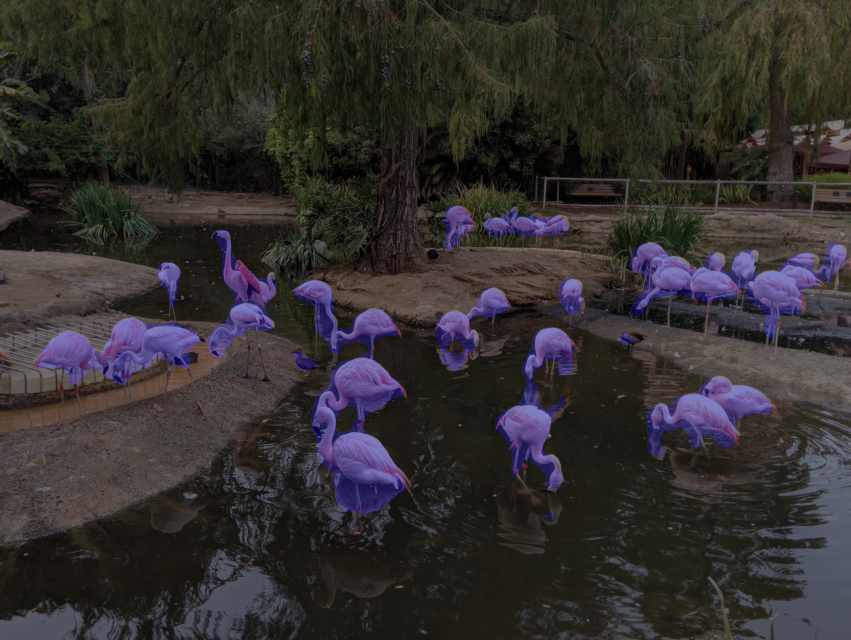} \\
         \includegraphics[width=0.25\linewidth]{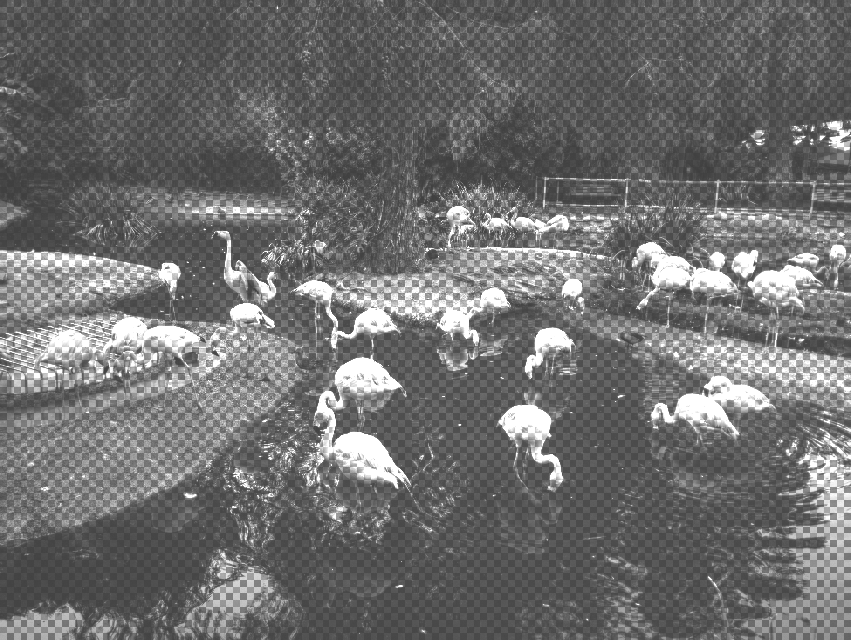} & \includegraphics[width=0.25\linewidth]{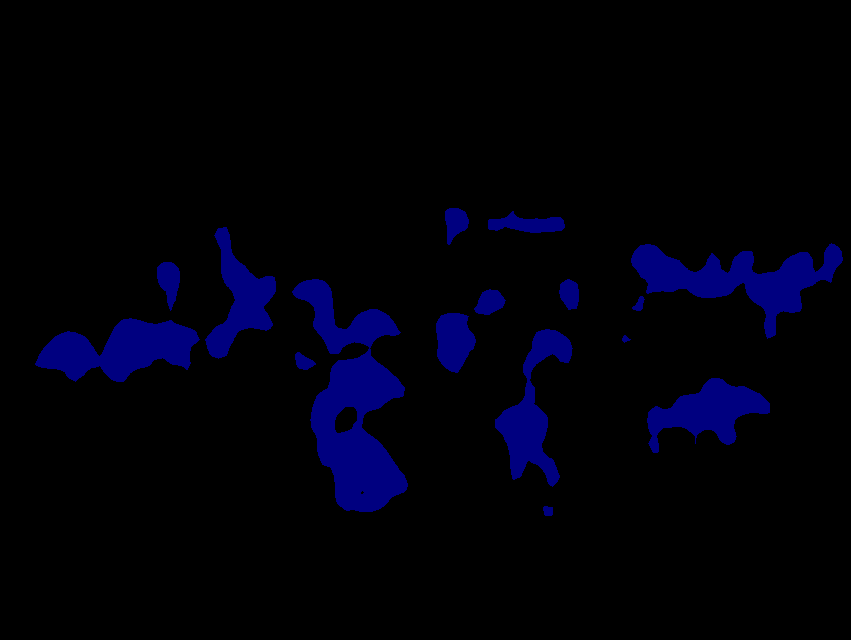} & \includegraphics[width=0.25\linewidth]{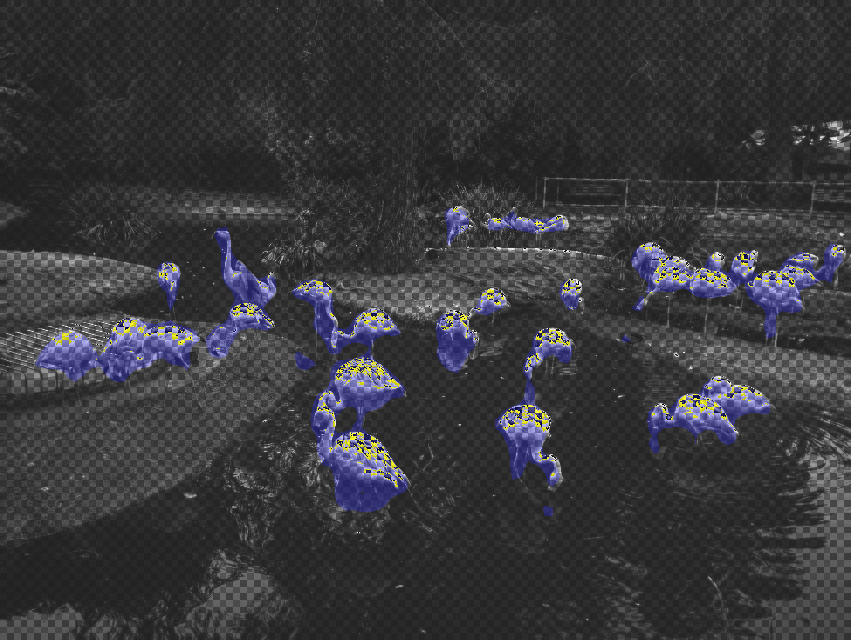} \\
         \includegraphics[width=0.25\linewidth]{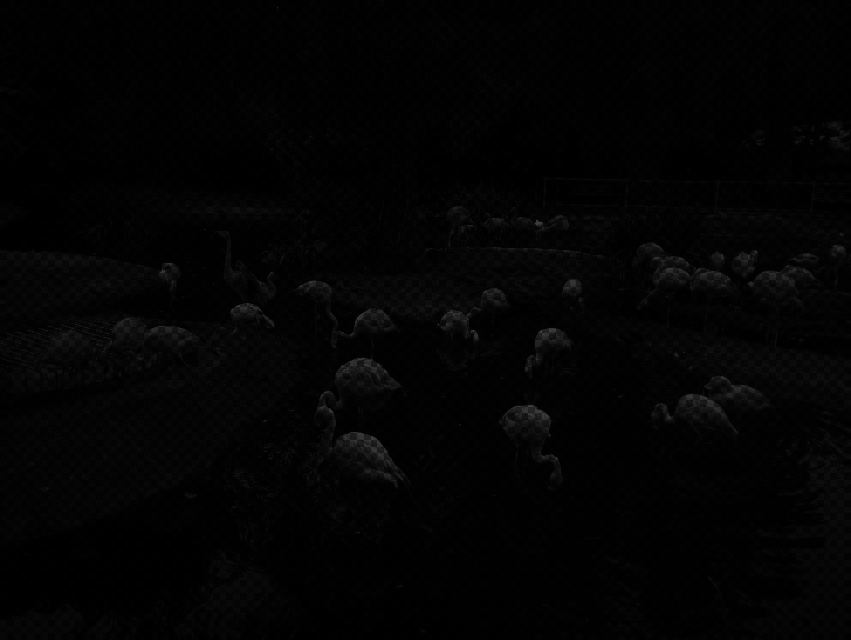} & \includegraphics[width=0.25\linewidth]{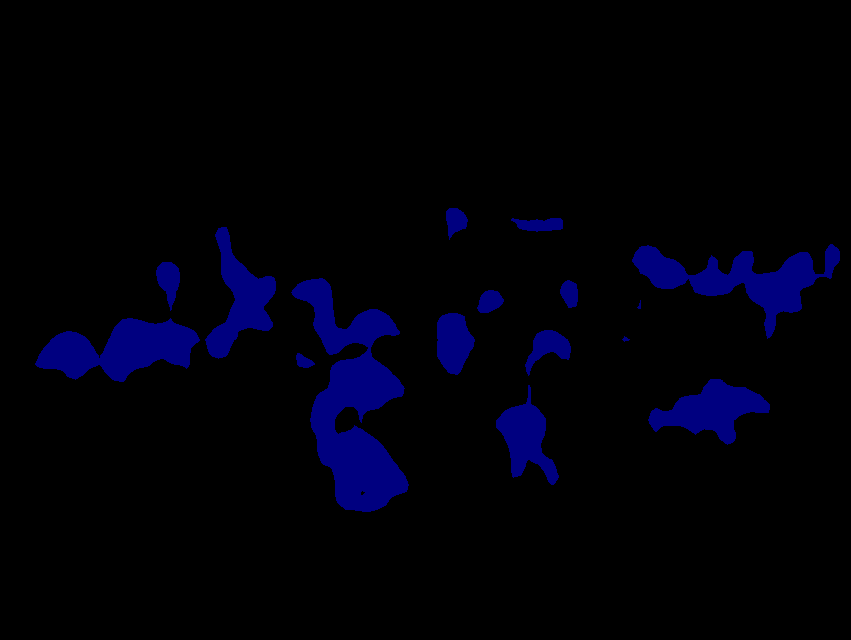} & \includegraphics[width=0.25\linewidth]{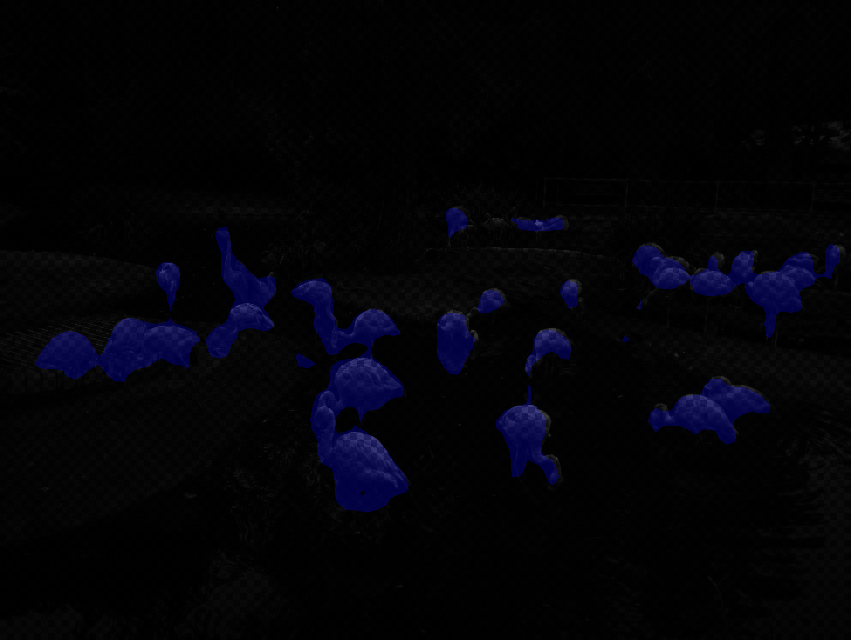} \\
         \\
         \includegraphics[width=0.25\linewidth]{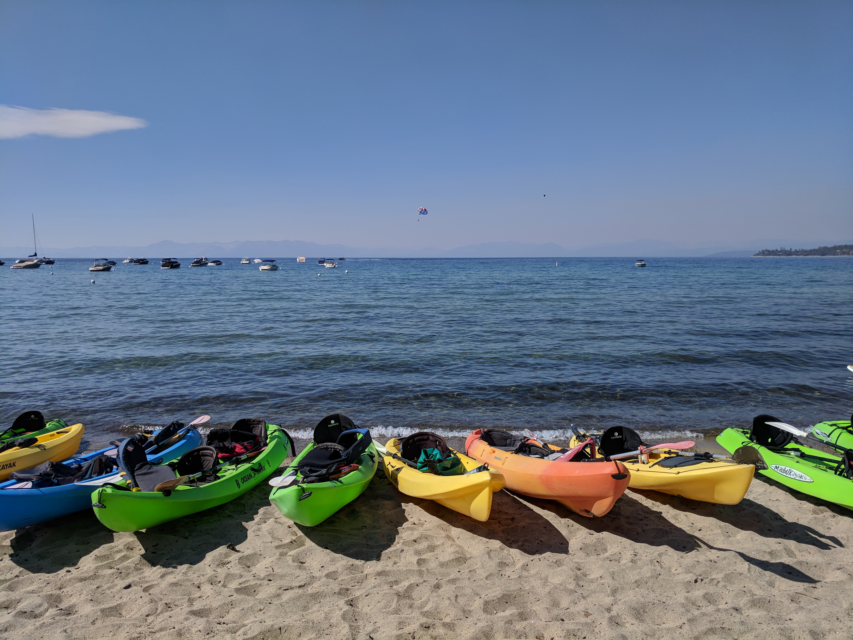} & \includegraphics[width=0.25\linewidth]{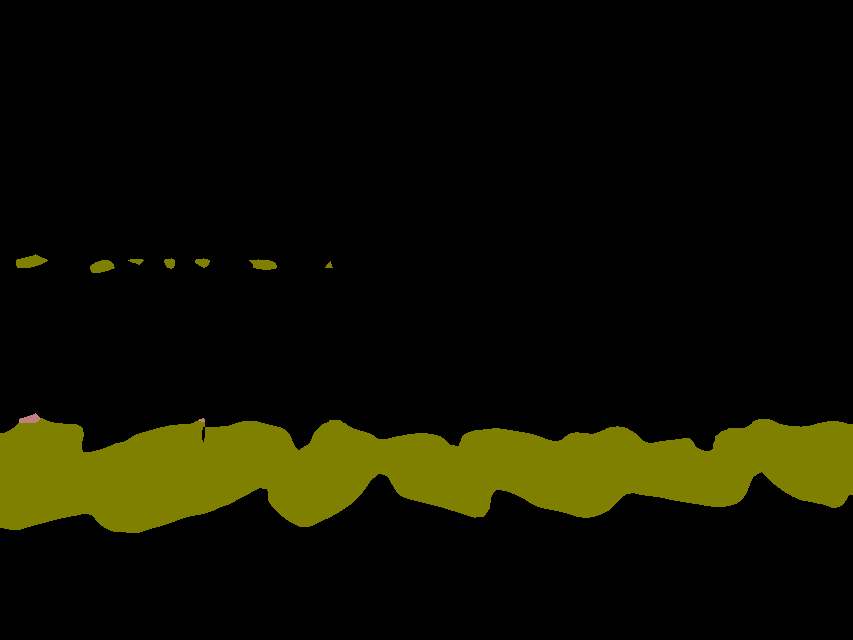} & \includegraphics[width=0.25\linewidth]{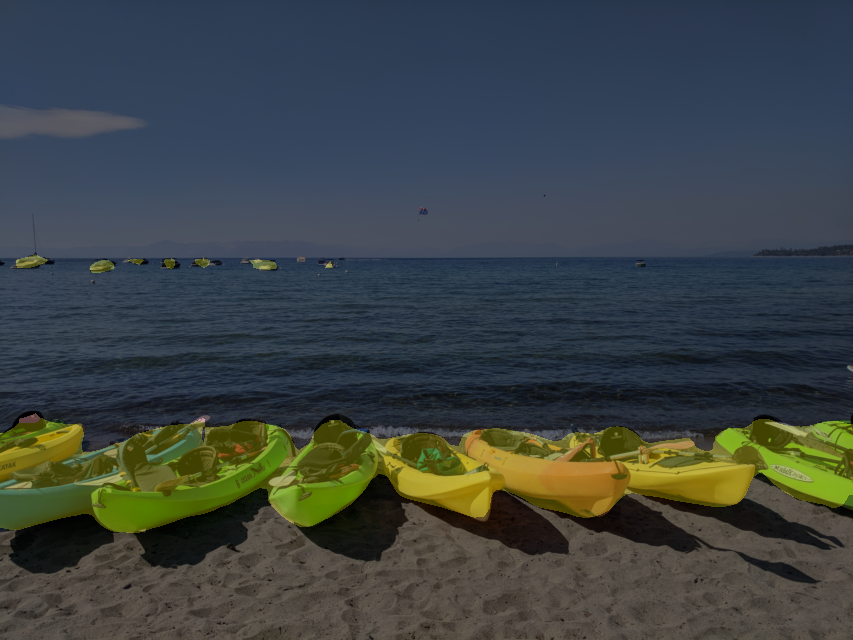} \\
         \includegraphics[width=0.25\linewidth]{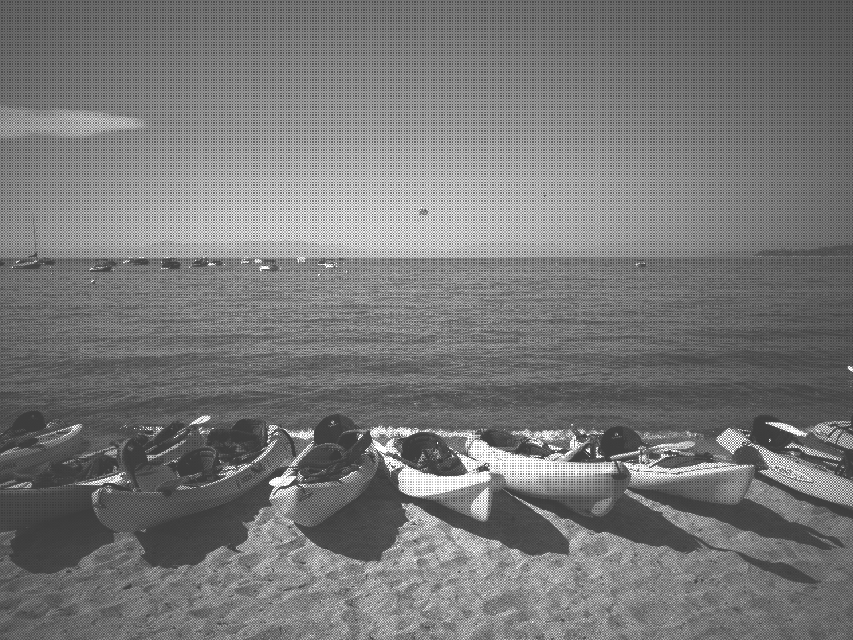} & \includegraphics[width=0.25\linewidth]{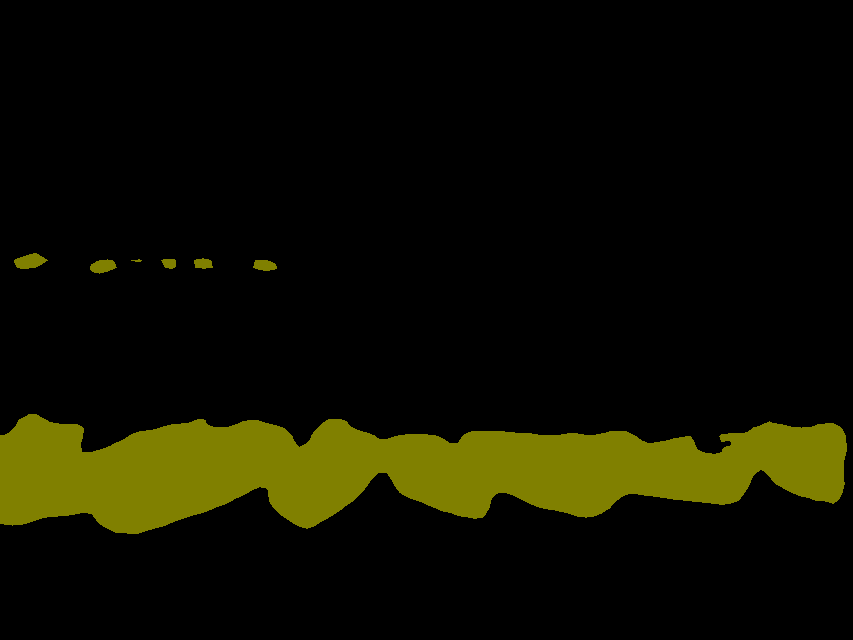} & \includegraphics[width=0.25\linewidth]{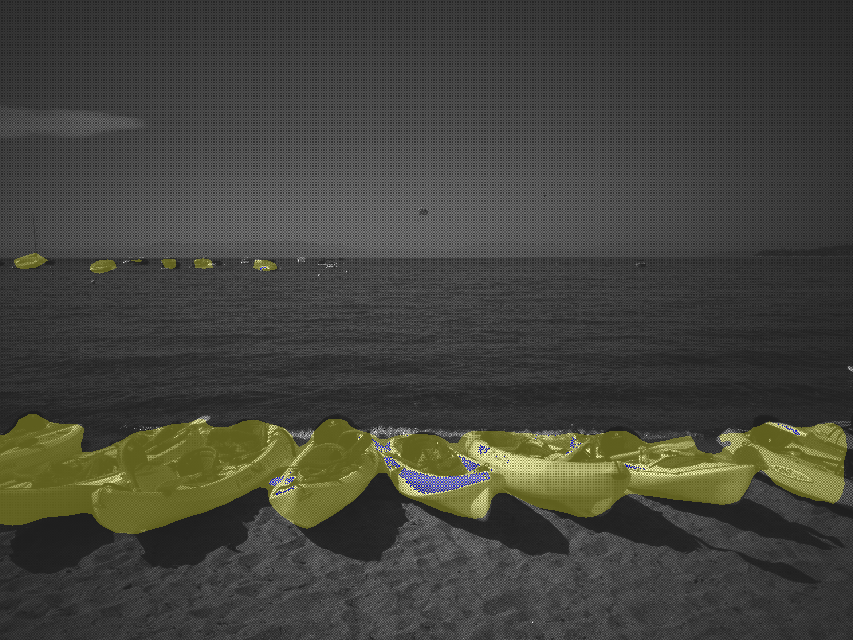} \\
         \includegraphics[width=0.25\linewidth]{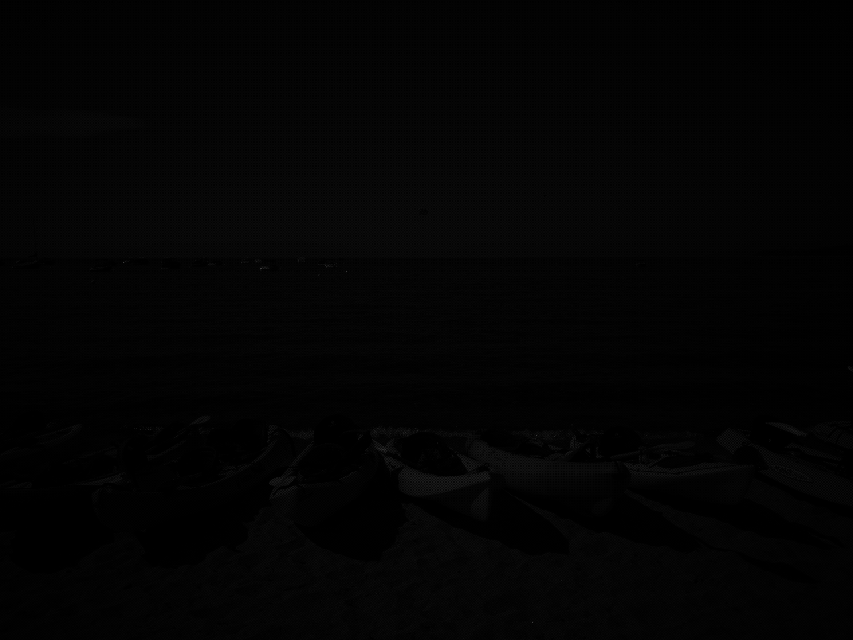} & \includegraphics[width=0.25\linewidth]{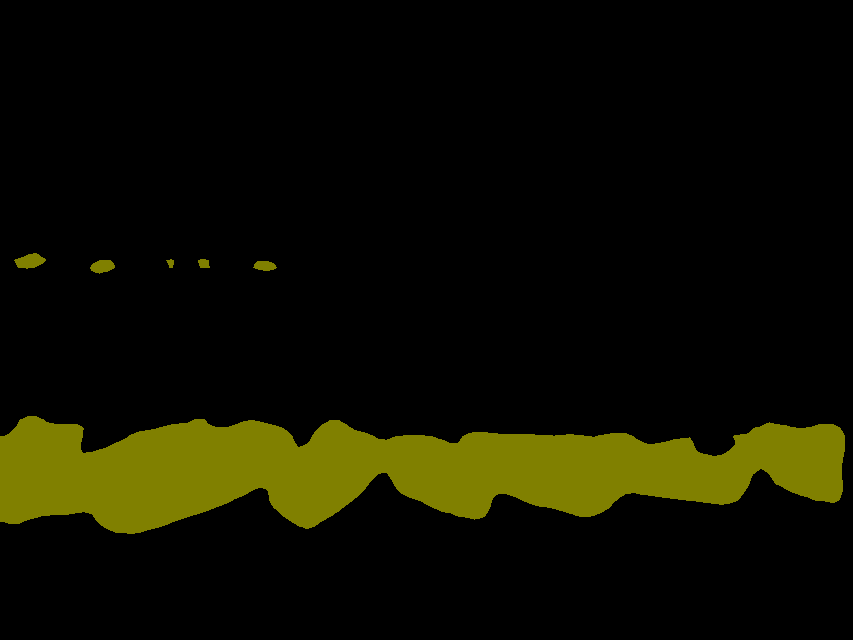} & \includegraphics[width=0.25\linewidth]{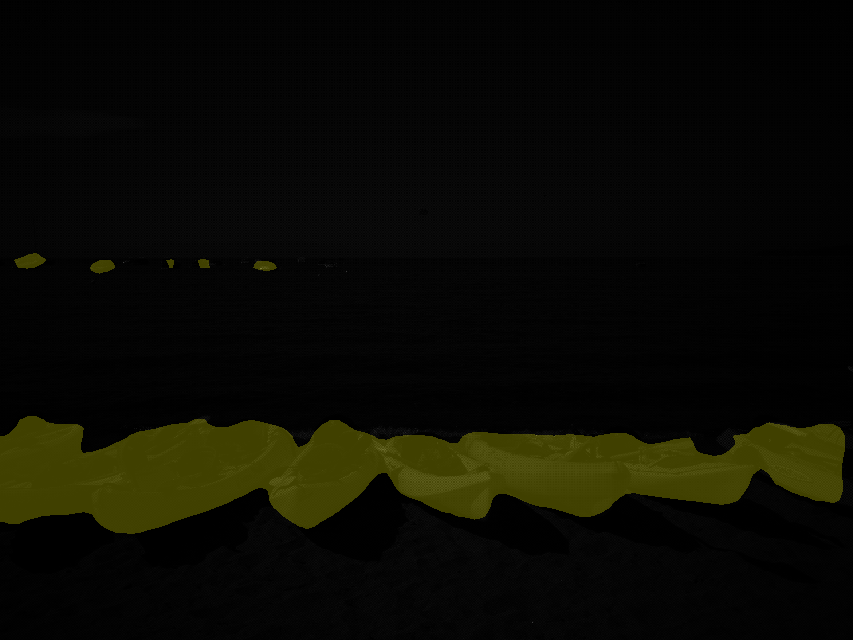} \\
    
    \end{tabular}
    \caption{\textbf{Additional semantic segmentation result examples.} For each example the top row is the ground truth RGB results; the middle row is our results for RAW input images; the bottom row is our results for short-exposure RAW images}
    \label{fig:seg2}
\end{figure*}

Figures \ref{fig:seg1} and \ref{fig:seg2} provide representative examples of the semantic segmentation results. To quantitatively assess the model performance we compute the Mean Intersection Over Union (mIOU) between the pretrained RGB model outputs and our RAW model (i.e. the agreement between the RGB outputs and the RAW outputs). The quantitative results are summarized in Table \ref{tab:segmentation}. We observed improvement over the training with labels baseline of $\sim 2.5\%$ mIOU for both RAW images and short-exposure RAW images. This is a significant improvement, but not as big as the improvement for classification. We suspect that the per-pixel output adds higher-level information providing better guidance to the student and thus the effect of having the soft-labels rather than the hard-labels is smaller.

\subsection{Ablation studies}

\begin{table}[h]
    \centering
    \normalsize	
    % \hspace{-10pt}
    \begin{tabular}{@{}lcccc@{}}
        \toprule
        & \multicolumn{2}{c}{ResNet18} & \multicolumn{2}{c}{MobileNetV2} \\
                                                    & top-1 & top-5 & top-1 & top-5 \\
         \midrule
         ISP Dist. (Ours) & \textbf{62.46} & \textbf{84.41} & \textbf{62.92} & \textbf{84.84} \\ 
         ISP Dist.-feat. loss & {61.35} & {83.70} & {61.43} & {83.95}\\
         ISP Dist.-feat. loss+ResNet50 & {62.43} & {84.18} & {62.62} & {84.14}\\ 
         \bottomrule
    \end{tabular}
% \vspace{-10pt}
    \caption{\textbf{Ablations.} Performed on the ImageNet experiment. Removing the feature loss results in a drop in performance. Using stronger teacher (ResNet50) improve performance.}
% \vspace{-15pt}
    \label{tab:l2-vs-strong-teacher}
\end{table}

\textbf{Partial model finetuning.}
Since we initialize our RAW model with ImageNet pre-trained weights we want to test how many of the layers need to be adapted to accommodate the RAW input. Is it just about local artifacts and finetuning only the first layers will be enough or there are global distortions that require higher-level features from deeper layers to adapt too? Table~\ref{tab:ablation} shows that indeed training just the first layers is beneficial, most of the gains are thanks to the finetuning of the first and second quarters of the network. Training the first half of ResNet18 is almost as good as training the full model.

\vspace{5pt}
\textbf{Features loss and stronger teacher.} Knowledge Distillation was originally used for distilling knowledge from stronger (more parameters, deeper) models to weaker ones. We verify performance can be further improved by using a stronger teacher in our case. We test the case where the teacher architecture is ResNet50 and the student is either ResNet18 or MobileNet V2. The teacher and student models' feature dimension is different, and even if it was the same since they are pretrained independently points in the two unaligned embedding spaces cannot be compared. For this reason, we drop the $\ell2$ loss on the features in this case. Table \ref{tab:l2-vs-strong-teacher} shows that dropping the features loss results in a drop in about $1\%$ drop in performance. Adding the stronger teacher improves the performance so it is comparable to the effect of features loss. In all our experiments we use the features loss. It might be possible to get even higher performance by combining the features loss with a stronger teacher (by having the same features dimension and enforcing embedding space alignment).

% \vspace{-20pt}
% \subsection{Real Data Detection}

\subsection{Things that did not work}
\ans{
For completeness, we also report ideas we hypothesized should improve performance but empirically did not.
}

\vspace{5pt}
\ans{
\textbf{Gradual blending.} We are trying to solve the domain adaptation problem from RGB to RAW. Unlike the usual domain adaptation task, in this case, we have pixel-aligned pairs of images from both domains. Since we have pixel alignment, we can generate infinitely many intermediate domains by interpolating between the RAW and RGB images, i.e. we can define a set of new domains $D_\alpha$ such that the images in $D_\alpha$ are $\{\alpha RGB_i + (1-\alpha)RAW_i\}$. This approach was used by \cite{arar2020focus} for the case of shifting from the classification of images where the background is masked out to the classification of the same images where the background is not masked.  We expected that gradually (linearly) shifting between the domains, i.e. first training the model to classify from domains with small \ans{$\alpha$} and finishing with $\alpha =1$, would help the model to learn better. But, in our experiments, we found the model finally converged to a similar performance as directly training for $\alpha =1$.
}

% \textbf{Gradual blending.} The problem we are trying to solve can be thought of as domain adaptation from the RGB image space to the RAW image space. Unlike the usual domain adaptation task, we enjoy the benefit of having pairs of images from the two domains with pixel-alignment. Since we have the pixel alignment, we can generate infinitely many intermediate domains by interpolating between the RAW and RGB image, i.e. we can define a set of new domains $D_\alpha$ such that the images in $D_\alpha$ are $\{\alpha RGB_i + (1-\alpha)RAW_i\}$. This approach was used by \cite{arar2020focus} for the case of shifting from classification of images where the background is masked out to classification of the same images where the background is not masked.  We expected that gradually (linearly) shifting between the domains, i.e. first training the model to classify from domains with small $alpha$ and finishing with $\alpha =1$, will help the model to learn better. But, in our experiments we found the model finally converge to similar performance as directly training for $\alpha =1$.

\vspace{5pt}
\textbf{Localized distillation.} We observed in the case of semantic segmentation that even training with hard labels yields very strong results. We hypothesized that the localized semantic signal helps the student learn better by providing more information. To test it we performed the experiment of distilling both the ISP and a classifier (similarly to Section \ref{sec:disc-isp}) but requiring spatial consistency between the teacher and student. To keep the spatial information we removed the global pooling layer and replaced the fully connected layer with a \ans{$1\times1$} convolution layer (with the same weights). We then applied the KD loss at each spatial location. However, in our experiments, this didn't improve the results beyond what we got with the `global' KD loss.

\section{Conclusions}
% \vspace{-10pt}
We have shown that it is possible to distill the knowledge of not just pre-trained models but also heuristically designed ISP, to improve the performance of classification and semantic segmentation models for RAW images. We have also shown improvement for short-exposure RAW, distilling the information of the physical process of acquiring a better signal. Our proposed ISP distillation is a step towards reaching similar performance on RAW images compared to RGB in high-level vision tasks. It can advance the deployment of vision models on RAW images in domains where the images are consumed by machines and not humans. This will save the computing cost of the ISP.

Possible future directions include: 
\begin{itemize}
    \item Combining ISP-Distillation with ISPs designed for vision tasks, i.e. distilling an ISP that was optimized for a vision task into the high-level model
    \item \ans{Exploring hybrid solutions where some blocks of the ISP are distilled into the high-level model and removed from the pre-processing pipeline while others (low compute) are kept}
    \item Extending ISP Distillation to other vision tasks (e.g. detection)
    \item Exploring using this method for other sensors that might better fit the needs of a machine vision system (e.g. event cameras)
\end{itemize}
% (i) 
% ֿ\ans{}
% (iii) Extending ISP Distillation to other vision tasks (e.g. detection); and (iii) Exploring using this method for other sensors that might better fit the needs of a machine vision system (e.g. event cameras).

\section*{Acknowledgment}
This work was supported by the European research council (ERC-StG 757497 PI Giryes).

% \clearpage
\bibliographystyle{ieee}
\bibliography{egbib}

\begin{thebibliography}{10}\itemsep=-1pt

\bibitem{arar2020focus}
M.~Arar, N.~Fish, D.~Daniel, E.~Tenetov, A.~Shamir, and A.~Bermano.
\newblock Focus-and-expand: Training guidance through gradual manipulation of
  input features.
\newblock {\em arXiv preprint arXiv:2007.07723}, 2020.

\bibitem{bagherinezhad2018label}
H.~Bagherinezhad, M.~Horton, M.~Rastegari, and A.~Farhadi.
\newblock Label refinery: Improving imagenet classification through label
  progression.
\newblock {\em arXiv preprint arXiv:1805.02641}, 2018.

\bibitem{borkar2019deepcorrect}
T.~S. Borkar and L.~J. Karam.
\newblock Deepcorrect: Correcting dnn models against image distortions.
\newblock {\em IEEE Transactions on Image Processing}, 28(12):6022--6034, 2019.

\bibitem{bucilua2006model}
C.~Bucilua, R.~Caruana, and A.~Niculescu-Mizil.
\newblock Model compression.
\newblock In {\em Proceedings of the 12th ACM SIGKDD International Conference
  on Knowledge Discovery and Data Mining}, pages 535--541, 2006.

\bibitem{buckler2017reconfiguring}
M.~Buckler, S.~Jayasuriya, and A.~Sampson.
\newblock Reconfiguring the imaging pipeline for computer vision.
\newblock In {\em Proceedings of the IEEE International Conference on Computer
  Vision}, pages 975--984, 2017.

\bibitem{chen2018learning}
C.~Chen, Q.~Chen, J.~Xu, and V.~Koltun.
\newblock Learning to see in the dark.
\newblock In {\em Proceedings of the IEEE Conference on Computer Vision and
  Pattern Recognition}, pages 3291--3300, 2018.

\bibitem{chen2017rethinking}
L.-C. Chen, G.~Papandreou, F.~Schroff, and H.~Adam.
\newblock Rethinking atrous convolution for semantic image segmentation.
\newblock {\em arXiv preprint arXiv:1706.05587}, 2017.

\bibitem{chi2020dynamic}
Y.~Chi, A.~Gnanasambandam, V.~Koltun, and S.~H. Chan.
\newblock Dynamic low-light imaging with quanta image sensors.
\newblock In {\em Computer Vision--ECCV 2020: 16th European Conference,
  Glasgow, UK, August 23--28, 2020, Proceedings, Part XXI 16}, pages 122--138.
  Springer, 2020.

\bibitem{diamond2017dirty}
S.~Diamond, V.~Sitzmann, S.~Boyd, G.~Wetzstein, and F.~Heide.
\newblock Dirty pixels: Optimizing image classification architectures for raw
  sensor data.
\newblock {\em arXiv preprint arXiv:1701.06487}, 2017.

\bibitem{Everingham10}
M.~Everingham, L.~Van~Gool, C.~K.~I. Williams, J.~Winn, and A.~Zisserman.
\newblock The pascal visual object classes (voc) challenge.
\newblock {\em International Journal of Computer Vision}, 88(2):303--338, June
  2010.

\bibitem{fukui2019distilling}
S.~Fukui, J.~Yu, and M.~Hashimoto.
\newblock Distilling knowledge for non-neural networks.
\newblock In {\em 2019 Asia-Pacific Signal and Information Processing
  Association Annual Summit and Conference (APSIPA ASC)}, pages 1411--1416.
  IEEE, 2019.

\bibitem{gnanasambandam2020image}
A.~Gnanasambandam and S.~H. Chan.
\newblock Image classification in the dark using quanta image sensors.
\newblock In {\em ECCV}, 2020.

\bibitem{hansen2019isp4ml}
P.~Hansen, A.~Vilkin, Y.~Khrustalev, J.~Imber, D.~Hanwell, M.~Mattina, and
  P.~N. Whatmough.
\newblock Isp4ml: Understanding the role of image signal processing in
  efficient deep learning vision systems.
\newblock {\em arXiv preprint arXiv:1911.07954}, 2019.

\bibitem{hasinoff2016burst}
S.~W. Hasinoff, D.~Sharlet, R.~Geiss, A.~Adams, J.~T. Barron, F.~Kainz,
  J.~Chen, and M.~Levoy.
\newblock Burst photography for high dynamic range and low-light imaging on
  mobile cameras.
\newblock {\em ACM Transactions on Graphics (Proc. SIGGRAPH Asia)}, 35(6),
  2016.

\bibitem{he2016deep}
K.~He, X.~Zhang, S.~Ren, and J.~Sun.
\newblock Deep residual learning for image recognition.
\newblock In {\em CVPR}, 2016.

\bibitem{hinton2015distilling}
G.~Hinton, O.~Vinyals, and J.~Dean.
\newblock Distilling the knowledge in a neural network.
\newblock {\em arXiv preprint arXiv:1503.02531}, 2015.

\bibitem{hong2021student}
G.~Hong, Z.~Mao, X.~Lin, and S.~H. Chan.
\newblock Student-teacher learning from clean inputs to noisy inputs.
\newblock In {\em Proceedings of the IEEE/CVF Conference on Computer Vision and
  Pattern Recognition}, pages 12075--12084, 2021.

\bibitem{8396981}
L.~J. Karam, T.~Borkar, Y.~Cao, and J.~Chae.
\newblock Generative sensing: Transforming unreliable sensor data for reliable
  recognition.
\newblock In {\em 2018 IEEE Conference on Multimedia Information Processing and
  Retrieval (MIPR)}, pages 100--105, 2018.

\bibitem{li2017learning}
Z.~Li and D.~Hoiem.
\newblock Learning without forgetting.
\newblock {\em IEEE transactions on pattern analysis and machine intelligence},
  40(12):2935--2947, 2017.

\bibitem{lin2014microsoft}
T.-Y. Lin, M.~Maire, S.~Belongie, J.~Hays, P.~Perona, D.~Ramanan,
  P.~Doll{\'a}r, and C.~L. Zitnick.
\newblock Microsoft coco: Common objects in context.
\newblock In {\em European conference on computer vision}, pages 740--755.
  Springer, 2014.

\bibitem{mosleh2020hardware}
A.~Mosleh, A.~Sharma, E.~Onzon, F.~Mannan, N.~Robidoux, and F.~Heide.
\newblock Hardware-in-the-loop end-to-end optimization of camera image
  processing pipelines.
\newblock In {\em Proceedings of the IEEE/CVF Conference on Computer Vision and
  Pattern Recognition}, pages 7529--7538, 2020.

\bibitem{peng2019few}
Z.~Peng, Z.~Li, J.~Zhang, Y.~Li, G.-J. Qi, and J.~Tang.
\newblock Few-shot image recognition with knowledge transfer.
\newblock In {\em Proceedings of the IEEE International Conference on Computer
  Vision}, pages 441--449, 2019.

\bibitem{romero2014fitnets}
A.~Romero, N.~Ballas, S.~E. Kahou, A.~Chassang, C.~Gatta, and Y.~Bengio.
\newblock Fitnets: Hints for thin deep nets.
\newblock {\em arXiv preprint arXiv:1412.6550}, 2014.

\bibitem{russakovsky2015imagenet}
O.~Russakovsky, J.~Deng, H.~Su, J.~Krause, S.~Satheesh, S.~Ma, Z.~Huang,
  A.~Karpathy, A.~Khosla, M.~Bernstein, et~al.
\newblock Imagenet large scale visual recognition challenge.
\newblock {\em International Journal of Computer Vision}, 115(3):211--252,
  2015.

\bibitem{sandler2018mobilenetv2}
M.~Sandler, A.~Howard, M.~Zhu, A.~Zhmoginov, and L.-C. Chen.
\newblock Mobilenetv2: Inverted residuals and linear bottlenecks.
\newblock In {\em CVPR}, 2018.

\bibitem{schwartz2018deepisp}
E.~Schwartz, R.~Giryes, and A.~M. Bronstein.
\newblock Deepisp: Toward learning an end-to-end image processing pipeline.
\newblock {\em IEEE TIP}, 2018.

\bibitem{sharma2018classification}
V.~Sharma, A.~Diba, D.~Neven, M.~S. Brown, L.~Van~Gool, and R.~Stiefelhagen.
\newblock Classification-driven dynamic image enhancement.
\newblock In {\em CVPR}, 2018.

\bibitem{tseng2019hyperparameter}
E.~Tseng, F.~Yu, Y.~Yang, F.~Mannan, K.~S. Arnaud, D.~Nowrouzezahrai, J.-F.
  Lalonde, and F.~Heide.
\newblock Hyperparameter optimization in black-box image processing using
  differentiable proxies.
\newblock {\em ACM Trans. Graph.}, 38(4):27--1, 2019.

\bibitem{wang2020deep}
Y.~Wang, Y.~Cao, Z.-J. Zha, J.~Zhang, and Z.~Xiong.
\newblock Deep degradation prior for low-quality image classification.
\newblock In {\em CVPR}, 2020.

\bibitem{wu2019visionisp}
C.-T. Wu, L.~F. Isikdogan, S.~Rao, B.~Nayak, T.~Gerasimow, A.~Sutic,
  L.~Ain-kedem, and G.~Michael.
\newblock Visionisp: Repurposing the image signal processor for computer vision
  applications.
\newblock In {\em ICIP}, 2019.

\bibitem{yahiaoui2019overview}
L.~Yahiaoui, J.~Horgan, B.~Deegan, S.~Yogamani, C.~Hughes, and P.~Denny.
\newblock Overview and empirical analysis of isp parameter tuning for visual
  perception in autonomous driving.
\newblock {\em Journal of Imaging}, 5(10):78, 2019.

\bibitem{zheng2016improving}
S.~Zheng, Y.~Song, T.~Leung, and I.~Goodfellow.
\newblock Improving the robustness of deep neural networks via stability
  training.
\newblock In {\em Proceedings of the ieee conference on computer vision and
  pattern recognition}, pages 4480--4488, 2016.

\bibitem{zheng2020optical}
Y.~Zheng, M.~Zhang, and F.~Lu.
\newblock Optical flow in the dark.
\newblock In {\em Proceedings of the IEEE/CVF Conference on Computer Vision and
  Pattern Recognition}, pages 6749--6757, 2020.

\end{thebibliography}

\begin{IEEEbiography}[{\includegraphics[width=1in,height=1.25in,clip,keepaspectratio]{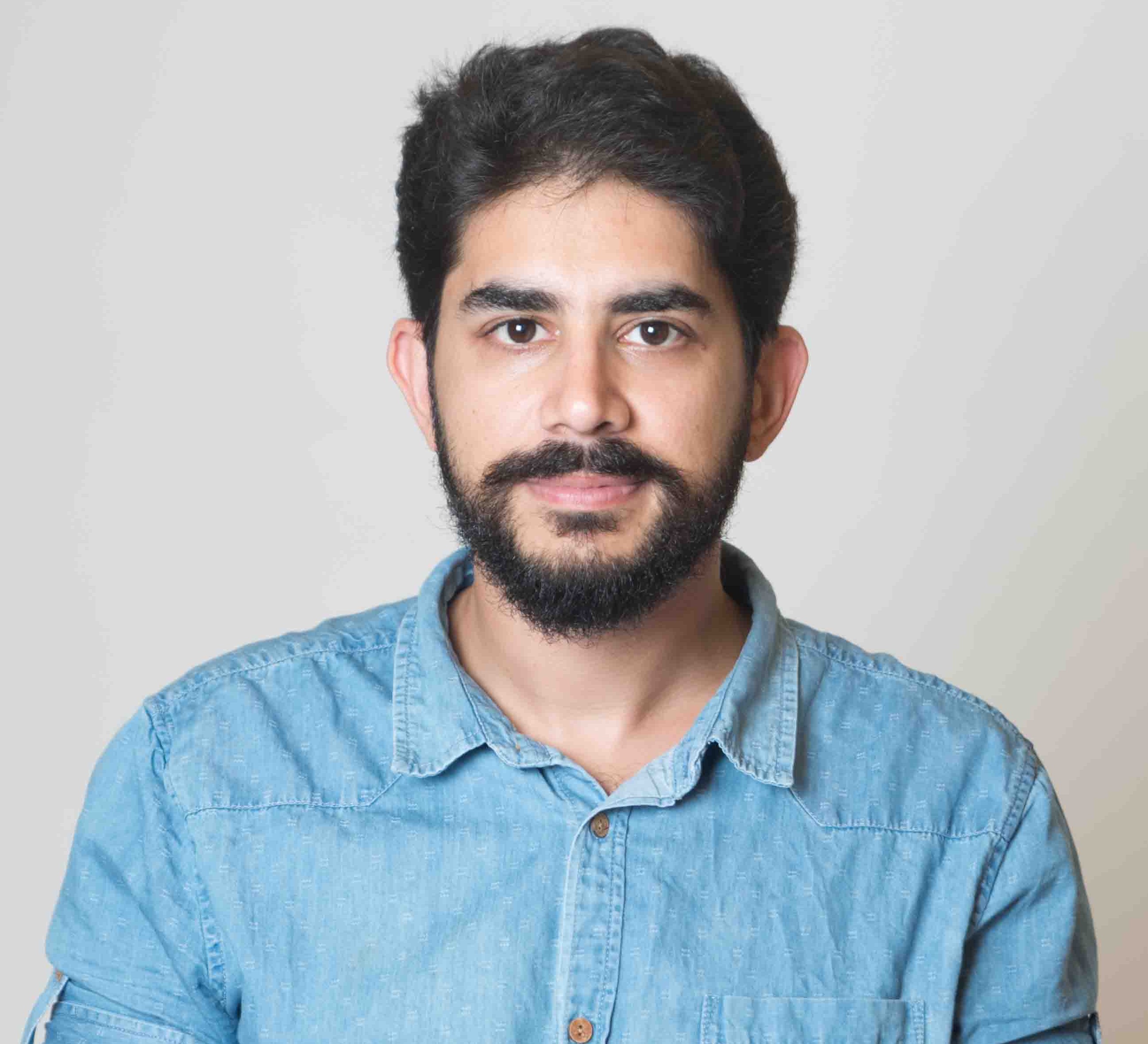}}]{Eli Schwartz}
is a Ph.D. Candidate in Electrical Engineering at Tel-Aviv University. Previously, he obtained an M.Sc in Electrical Engineering from Tel Aviv University (2018) and a B.Sc. in Electrical Engineering from the Technion - Israel Institute of Technology (2013). 
Eli has authored 17 academic papers with publications in top journals and conferences and served as a program chair for the Learning with Limitted Labels Workshop at CVPR.
He is the recipient of IBM Fellowship (2020).

Eli is currently a Research Scientist working for IBM Research.
Previously, Eli co-founded and served as the CTO of a robotics startup - Inka Robotics that developed the world's first tattooing robot (2015-2017). Before that, Eli was with Microsoft, developing algorithms for the HoloLens AR headset (2013-2016). Eli has also worked in the chip design industry for Qualcomm (2011-2013) and IBM (2008-2011).
\end{IEEEbiography}

\begin{IEEEbiography}[{\includegraphics[width=1in,height=1.25in,clip,keepaspectratio]{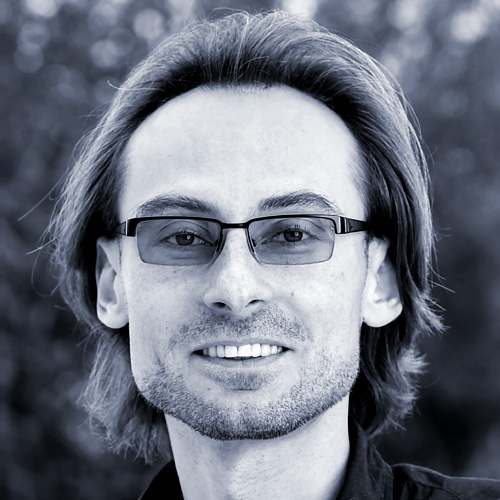}}]{Alex Bronstein}
(Fellow, IEEE) is currently a Professor of computer
science with the Technion-Israel Institute of Technology and a
Principal Engineer with Intel Corporation. His research interests
include numerical geometry, computer vision, and machine learning.
He has authored over 100 publications in leading journals and
conferences, over 30 patents and patent applications, the research
monograph Numerical Geometry of Non-Rigid Shapes, and edited
several books. Highlights of his research were featured in CNN, SIAM
News, and Wired. In addition to his academic activity, he co-founded
and served as the Vice President of technology in the Silicon Valley
start-up company Novafora, from 2005 to 2009. He was a Co-Founder
and one of the main inventors and developers of the 3D sensing
technology with the Israeli startup Invision, subsequently acquired
by Intel, in 2012. His technology is currently the core of the Intel
RealSense 3D camera integrated into a variety of consumer electronic
products. He is also a Co-Founder of the Israeli video search startup
Videocites and the London-based startup Sibylla, where he serves as
the Chief Scientist. He is a Fellow of the IEEE for his contribution to
3D imaging and geometry processing.
\end{IEEEbiography}

\begin{IEEEbiography}[{\includegraphics[width=1in,height=1.25in,clip,keepaspectratio]{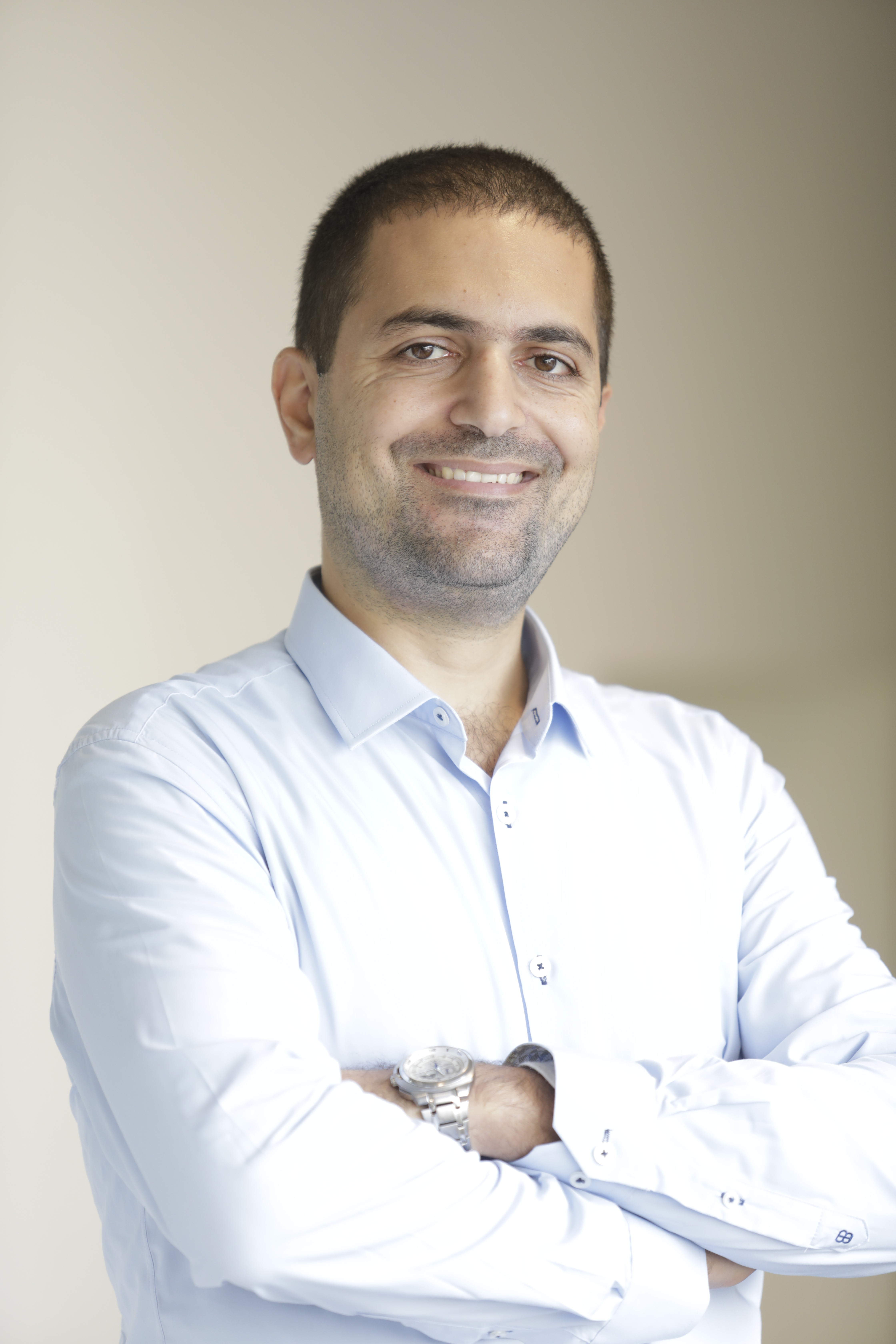}}]{Raja Giryes}
\textbf{\emph{Raja Giryes}} (raja@tauex.tau.ac.il) is an Associate Professor in the school of electrical engineering at Tel Aviv University. His research interests lie at the intersection between signal and image processing and machine learning, and in particular, in deep learning, inverse problems, sparse representations, computational photography, and signal and image modeling. Raja received the EURASIP best P.hD. award, the ERC-StG grant, Maof prize for excellent young faculty (2016-2019), VATAT scholarship for excellent postdoctoral fellows (2014-2015), Intel Research and Excellence Award (2005, 2013), the Excellence in Signal Processing Award (ESPA) from Texas Instruments (2008) and was part of the Azrieli Fellows program (2010-2013). He is an associate editor in IEEE Transactions on Image Processing and Elsevier Pattern Recognition and has organized workshops and tutorials on deep learning theory in various conferences including ICML, CVPR, ECCV and ICCV. He is also a co-organizer of the Israel computer vision day. He is an IEEE Senior Member and a member of the Israeli Young Academy since 2022.
\end{IEEEbiography}

\end{document}